\documentclass[10pt,twocolumn,letterpaper]{article}

\usepackage[pagenumbers]{iccv} % To force page numbers, e.g. for an arXiv version

\usepackage{times}
\usepackage{epsfig}
\usepackage{graphicx}
\usepackage{amsmath}
\usepackage{amssymb}

\usepackage{multirow}
\usepackage{tabularx}
\usepackage{booktabs}
\usepackage{longtable}
\usepackage{makecell}
\usepackage{afterpage}

\usepackage[accsupp]{axessibility}  % Improves PDF readability for those with disabilities.
\usepackage[belowskip=-5pt,aboveskip=3pt,font=footnotesize,labelfont=bf]{caption} % re-arrange tables on last page, select
\usepackage{dsfont}
\usepackage{pifont}
\usepackage{url}
\usepackage{color}
\usepackage[export]{adjustbox}
\usepackage{comment}
\usepackage{algorithm} 
\usepackage{multirow}
\usepackage{diagbox}

\usepackage[utf8]{inputenc} % allow utf-8 input
\usepackage[T1]{fontenc}    % use 8-bit T1 fonts
\usepackage{url}            % simple URL typesetting
\usepackage{amsfonts}       % blackboard math symbols
\usepackage{nicefrac}       % compact symbols for 1/2, etc.
\usepackage{microtype}      % microtypography
\usepackage{enumitem}
\usepackage{blindtext}% for dummy text only
\usepackage{sidecap}
\usepackage{wrapfig}
\usepackage[most]{tcolorbox}
\tcbuselibrary{skins, breakable, theorems}
\usepackage{tikz}
\usetikzlibrary{fadings}
\usepackage{dsfont}

\usepackage[dvipsnames]{xcolor}
\definecolor{redmark}{rgb}{0.8, 0.0, 0.0}
\definecolor{greenmark}{rgb}{0.0, 0.5, 0.0}

\definecolor{dg}{rgb}{0,0.694,0.298}
\definecolor{purple}{rgb}{0.4,0.176,0.569}
\definecolor{royalblue}{RGB}{65,105,225}
\usepackage{pifont}% http://ctan.org/pkg/pifont
\newcommand{\figref}[1]{Fig.~\ref{#1}}

\newcommand{\secref}[1]{Sec.~\ref{#1}}
\newcommand{\tableref}[1]{Tab.~\ref{#1}}

\makeatletter
\DeclareRobustCommand\onedot{\futurelet\@let@token\@onedot}
\def\@onedot{\ifx\@let@token.\else.\null\fi\xspace}
\def\eg{\emph{e.g}\onedot} 
\def\ie{\emph{i.e}\onedot}

\makeatother

\definecolor{americanrose}{rgb}{1.0, 0.01, 0.24}

\definecolor{mypink}{RGB}{255, 234, 229}
\definecolor{myblue}{RGB}{220, 234, 247}
\definecolor{mygray}{RGB}{217, 217, 217}
\definecolor{techgreen}{RGB}{218, 242, 209}
\definecolor{techblue}{RGB}{219, 227, 243}

\definecolor{rebuttalred}{RGB}{255,200,200}
\definecolor{rebuttalgreen}{RGB}{200,255,200}

\definecolor{iccvblue}{rgb}{0.21,0.49,0.74}
\usepackage[pagebackref,breaklinks,colorlinks,allcolors=iccvblue]{hyperref}

\def\paperID{4770} % *** Enter the Paper ID here
\def\confName{ICCV}
\def\confYear{2025}
\title{\textsc{MAGIC}: Mastering Physical Adversarial Generation in Context through Collaborative LLM Agents}

\author{Yun Xing$^{1,3}$\thanks{Work done during internship at CFAR \& IHPC, A*STAR. $^\dagger$Co-corresponding author, email address: tsingqguo@ieee.org.
}
, Nhat Chung$^{1,5*}$, Jie Zhang$^{1}$, Yue Cao$^{1,2}$, Ivor Tsang$^{1,2}$,  Yang Liu$^{2}$, 
Lei Ma$^{3,4}$,  Qing Guo$^{1\dagger}$
\\
$^1$ CFAR and IHPC, Agency for Science, Technology and Research (A*STAR), Singapore \\
$^2$ Nanyang Technological University, Singapore
\quad $^3$ University of Alberta, Canada \\
\quad $^4$ The University of Tokyo, Japan
\quad $^5$ VNU-HCM, Vietnam
}

\begin{document}

\let\tmptwocolumn\twocolumn
\renewcommand{\twocolumn}[1][]{%
    \tmptwocolumn[{#1}{
    \begin{center}
        \includegraphics[width=\textwidth]{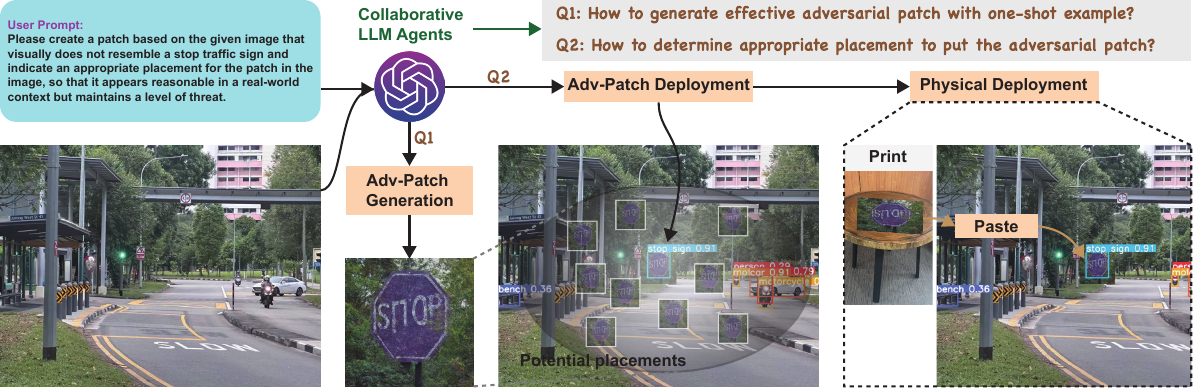}
        \vspace{-10pt}
        \captionsetup{type=figure}
        \captionof{figure}{Intuitive idea of our proposed \textsc{MAGIC} framework. Given the user prompt indicates the attack subject and objective and an image of a real-world scene, our method aims to generate an attack-effective adversarial patch and automatically deploy it into the physical scene with naturalness. We propose to realize the goals through multi-modal agent planning and leverage multiple LLM-based agents to achieve collaborative generation and deployment.}
        \label{fig:cmp}
    \end{center}
    }]
}

\maketitle

\begin{abstract}
Physical adversarial attacks in driving scenarios can expose critical vulnerabilities in visual perception models. However, developing such attacks remains challenging due to diverse real-world environments and the requirement for maintaining visual naturality.
Building upon this challenge, we reformulate physical adversarial attacks as a one-shot patch generation problem. Our approach generates adversarial patches through a deep generative model that considers the specific scene context, enabling direct physical deployment in matching environments.
The primary challenge lies in simultaneously achieving two objectives: generating adversarial patches that effectively mislead object detection systems while determining contextually appropriate deployment within the scene.
We propose MAGIC (Mastering Physical Adversarial Generation In Context), a novel framework powered by multi-modal LLM agents to address these challenges. MAGIC automatically understands scene context and generates adversarial patch through the synergistic interaction of language and vision capabilities.
In particular, MAGIC orchestrates three specialized LLM agents: The adv-patch generation agent (GAgent) masters the creation of deceptive patches through strategic prompt engineering for text-to-image models. The adv-patch deployment agent (DAgent) ensures contextual coherence by determining optimal deployment strategies based on scene understanding. The self-examination agent (EAgent) completes this trilogy by providing critical oversight and iterative refinement of both processes.
We validate our method on both digital and physical levels, \ie, nuImage and manually captured real-world scenes, where both statistical and visual results prove that our MAGIC is powerful and effective for attacking widely applied object detection systems, \ie, YOLO and DETR series.
\end{abstract}

\vspace{-25pt}
\section{Introduction}
\label{sec:intro}
% Comment
% Make our story be like an extension of a previous work for attack about the CVPR paper: https://arxiv.org/abs/2308.15692
% Focus on generating attacks against detection,
% Detection, making the object disappear, making the object appear
% instruction-based natural attacks on detection
% we do the attacks digitally and physically
%
% for reasoning and attack generation

% physical adversarial attacks can expose vulnerabilities
Adversarial attacks serve as a crucial method for evaluating the robustness and safety of deep learning models. Physical adversarial attacks, in particular, focus on creating adversarial patches that can be printed and deployed in the real world to mislead visual perception models \cite{trafficSignAttackSurvey2023, robustPhysicalAttack2018, physicalAdvAttackSurvey2024, NDDA2024cvpr, cao2024scenetap}.
Unlike their digital counterparts, physical adversarial attacks can directly expose vulnerabilities in deployed models during practical application \cite{NDDA2024cvpr}.
% physical adversarial attacks can be deployed directly in real-world settings, exposing vulnerabilities in deep learning models during practical application \cite{NDDA2024cvpr}.
%
This is crucial for security-sensitive applications such as autonomous driving (AD) \cite{ADReview2024TPAMI, ADReview2024ESA, nuScenes2020data}. When object detectors widely used in AD misidentify critical traffic participants \cite{objectDetection3DReview2023, nuScenes2020data, Jia2023FastAA, deruyttere2019talk2car}, \eg, traffic signs, they may trigger incorrect vehicle responses, potentially creating hazardous situations for both drivers and pedestrians such as sudden stops on highway. 
By revealing these weaknesses, physical adversarial attacks underscore the need for robust perception systems in AD.
% Physical adversarial attacks can expose these vulnerabilities, highlighting the importance of robust perception systems in AD scenarios.

As shown in \figref{fig:cmp}, physical adversarial attacks face two primary challenges: \ding{182} ensuring patch effectiveness against diverse real-world factors (\eg, color shifts, scale variation, view angle changes), and \ding{183} preserving patch stealthiness within the scene, that is fooling deep models without drawing human attention.
Traditionally, optimization-based adversarial patch generation methods struggle to achieve robust performance against real-world factors and state-of-the-art (SOTA) object detectors.
The recent work \cite{NDDA2024cvpr} demonstrated that diffusion models can generate robust and stealthy adversarial patches via prompt engineering without optimization.
However, this approach overlooks the scene context in the deployment environment, compromising both attack effectiveness and stealthiness.
Specifically, \cite{NDDA2024cvpr} fails to consider three critical aspects: \ding{182} the influence of the deployment process on patch's attack performance, \ding{183} the patch's ability to remain effective and stealthy within the scene, and \ding{184} the suitable deployment setup of the patch within the scene.
In \secref{sec:revisit}, we present comprehensive analyses demonstrating that scene context and the deployment strategy significantly influence the performance of adversarial patches.

These challenges arise from the insufficient scene context understanding and the inability to integrate it into both the adversarial patch generation and deployment.
A single deep model is inadequate for addressing these challenges, as scene understanding, patch generation, and deployment are distinct tasks requiring specialized capabilities.
To address these challenges, we reformulate physical adversarial attacks as an one-shot patch generation problem, and develop a framework that automatically generates and deploys adversarial patches, particularly factoring in the target environment image with user-specified attack objectives.
To achieve this, we propose \textsc{MAGIC} \textit{(Master physical Adversarial Generation In Context through collaborative multi-modal LLM agents)}, a framework where multiple LLM agents collaboratively generate and deploy effective, stealthy adversarial patches based on comprehensive scene analyses.
Our MAGIC operates in three key stages-(i) patch proposal, (ii) patch deployment, and (iii) patch refinement-powered by three specialized LLM agents (GAgent, DAgent, and EAgent) to master physical adversarial attack in real-world AD settings.
In summary, our contributions are summarized as follows,
\begin{itemize}

    \item We investigate the effectiveness of diffusion-based physical adversarial attacks \cite{NDDA2024cvpr} across diverse environments and deployment strategy, revealing that attack efficacy is highly dependent on scene context and patch deployment.

    \item We propose MAGIC, a novel framework that leverages collaborative multi-modal LLM agents to determine both physical adversarial patch generation and deployment based on comprehensive scene analyses. To the best of our knowledge, this is the first approach that combines LLM agents with physical adversarial generation.

    \item We evaluate MAGIC by attacking three well-known object detectors (YOLOv5, RT-DETR, YOLOv10) using both synthetic (\ie, digital patch insertion) and physical (\ie, real-world patch capturing) approaches, providing extensive analysis of their vulnerabilities in driving scenarios.

\end{itemize}

% \qing{working here.}

% To the best of our knowledge, ours is also the first study to improve and extend NDDA into the physical world. 

% figure: show the detection confidence of our patch and NDDA patch,

% \begin{figure}
%     \centering
%     \includegraphics[width=0.5\linewidth]{}
%     \caption{Caption}
%     \label{fig:enter-label}
% \end{figure}

\section{Related Works}
\label{sec:relate}

\noindent \textbf{Physical Adversarial Attacks.}
Physical adversarial attacks have been designed to stress-test perception systems used for AD in the real world~\cite{physical4ad2023}, particularly against classification~\cite{brown2017adversarial, casper2022robust, doan2022tnt, eykholt2018robust, zhong2022shadows}, detection~\cite{zhang2019camou, suryanto2022dta, yang2018building, thys2019fooling, xu2020adversarial, chen2018shapeshifter}, and among other applications ~\cite{cheng2022physical, kong2020physgan, ding2021towards,chung2024towards,guo2020spark}. These attacks manipulate models to misclassify or overlook the targets~\cite{ADReview2024TPAMI}, revealing practical threats to their deployment. In particular, physical adversarial patches~\cite{zhang2023capatch, wei2023unified, du2022physical, wei2022simultaneously, doan2022tnt, NDDA2024cvpr} are easily replicable and widely used to induce mispredictions in deployed models. Consequently, research on physical adversarial attacks, including attack effectiveness~\cite{brown2017adversarial,thys2019fooling,wei2023unified} and attack stealthiness~\cite{liu2019perceptual,wang2021dual,tan2021legitimate, huang2022ala}, remains essential for informing the development of safety-critical systems.
% Inspired by the potential of natural denoising diffusion attack~\cite{NDDA2024cvpr} and its lack of comprehensive analyses in the physical world, we propose a one-shot physical patch generation framework for object detectors using multi-agent reasoning.

\noindent \textbf{Adversarial Attack Design.}
Adversarial attacks exploit neural network vulnerabilities by misleading model predictions~\cite{goodfellow2015explainadv, segPGD2022, sensen2024multimodal, ad2022sok, MetaRepair2024}, which can diverge from human judgment to raise alarming safety concerns~\cite{physical4ad2023}. Conventionally, existing works add subtle perturbations that are not noticeable to humans~\cite{goodfellow2015explainadv, segPGD2022, sensen2024multimodal}, put adversarial stickers~\cite{stickerattack2018} or place small patches to the attack scenarios~\cite{brown2017adversarial, diffpgd2023, natpatch2021, wang2021universal, doan2022tnt, liu2020bias}. Hendrycks et al.~\cite{natadvex2021} found that even unedited images in the wild can pose adversarial effects. It has been hypothesized that these adversarial attacks are caused by non-robust, target-specific features that are incomprehensible to humans, rather than by inherent model issues~\cite{nonrobust2019, NDDA2024cvpr, natadvex2021}. Thus, a recent study~\cite{NDDA2024cvpr} employs diffusion models to generate natural adversarial examples through intentional text prompt alteration to attack object detectors. 
% However, its approach relies on manual prompting to create a static dataset of limited relevance to real-world AD scenarios.  In contrast, we explore how natural adversarial patterns can be automatically generated and dynamically evaluated in physical environments.

\noindent \textbf{Multimodal LLM Reasoning.}
Inspired by the emergent capabilities of LLMs in key techniques such as zero-shot prompting~\cite{cot2022}, in-context reasoning~\cite{incontext2024}, multi-modal reasoning~\cite{LLaVA-Grounding2024, vila2024}, and self-feedback~\cite{innermonologue2022}, autonomous agents have made significant strides in mimicking human interactions~\cite{saycan2022, multiagentdebate2024, gpt4drive2023, autogpt2024, yolas2024, mtid2024}. While language-based agents~\cite{incontext2024,roleplay2023} pioneered such interactions in text-based contexts, multi-modal embodied systems~\cite{saycan2022, gpt4drive2023, autogpt2024, yolas2024, mtid2024, mp52024} have extended these capabilities to real-world scenarios by integrating image~\cite{mtid2024, wang2024segllm, blink2024}, video~\cite{gpt4drive2023, shen2024longvu} and audio~\cite{autogpt2024}. Notably,~\cite{mtid2024} incorporated LLM reasoning for fine-grained diffusion results, demonstrating the potential of self-feedback in improving outcomes. 
% Unlike prior work, we integrate multi-modal LLM reasoning with diffusion to emulate human adversaries in real-world settings. By leveraging LLM reasoning and feedback, our framework offers controllable and dynamic physical attack generation against visual detectors for AD.

\section{Revisit Natural Denoising Diffusion Attack}
\label{sec:revisit}
% 1. NDDA is the pioneer work, 2. NDDA sets the baseline, 3. NDDA's deficiency

Our study is inspired by the practical obstacles encountered during real-world deployment of Natural Denoising Diffusion Attack (NDDA) \cite{NDDA2024cvpr}.
Thus, we first briefly revisit NDDA then analyze the problems we encountered during the deployment which motivated us to propose the MAGIC.

% %
% \begin{figure}[t]
%     \centering
%     \includegraphics[width=\linewidth]{figs/rq2.pdf}
%     \caption{Research question 2.}
%     \label{fig:rq2}
% \end{figure}
% %

% %
% \begin{figure}[t]
%     \centering
%     \includegraphics[width=\linewidth]{figs/rq34.pdf}
%     \caption{Research question 3 and 4.}
%     \label{fig:rq34}
% \end{figure}
% %

\subsection{Preliminaries of NDDA}
\label{ssec:prelim}

NDDA leverages SOTA text-to-image (T2I) models, \eg, Stable Diffusion \cite{rombach2022high}, to generate physical adversarial patches through strategic manipulation of the text prompts.
Given a text prompt $\mathcal{T}$ that specifies both a subject (\eg, stop sign) and related robust features \cite{ge2022contributions, grill2004human} (\ie, shape, color, text, and pattern), the diffusion model generates a corresponding patch.
NDDA then systematically modifies the patch generation by altering the prompt $\mathcal{T}$, which selectively removes robust features from the normal stop sign.
As reported in \cite{NDDA2024cvpr}, the selective removal of robust features yields patches that could achieve high error detection rates while remaining stealthy to human observers.
Such adversarial attack effectiveness aligns with the findings from \cite{ilyas2019adversarial}, which demonstrate adversarial attacks that exploit non-robust features are predictive for DNNs but incomprehensible to humans.

\begin{figure}[t]
    \centering
    \includegraphics[width=\linewidth]{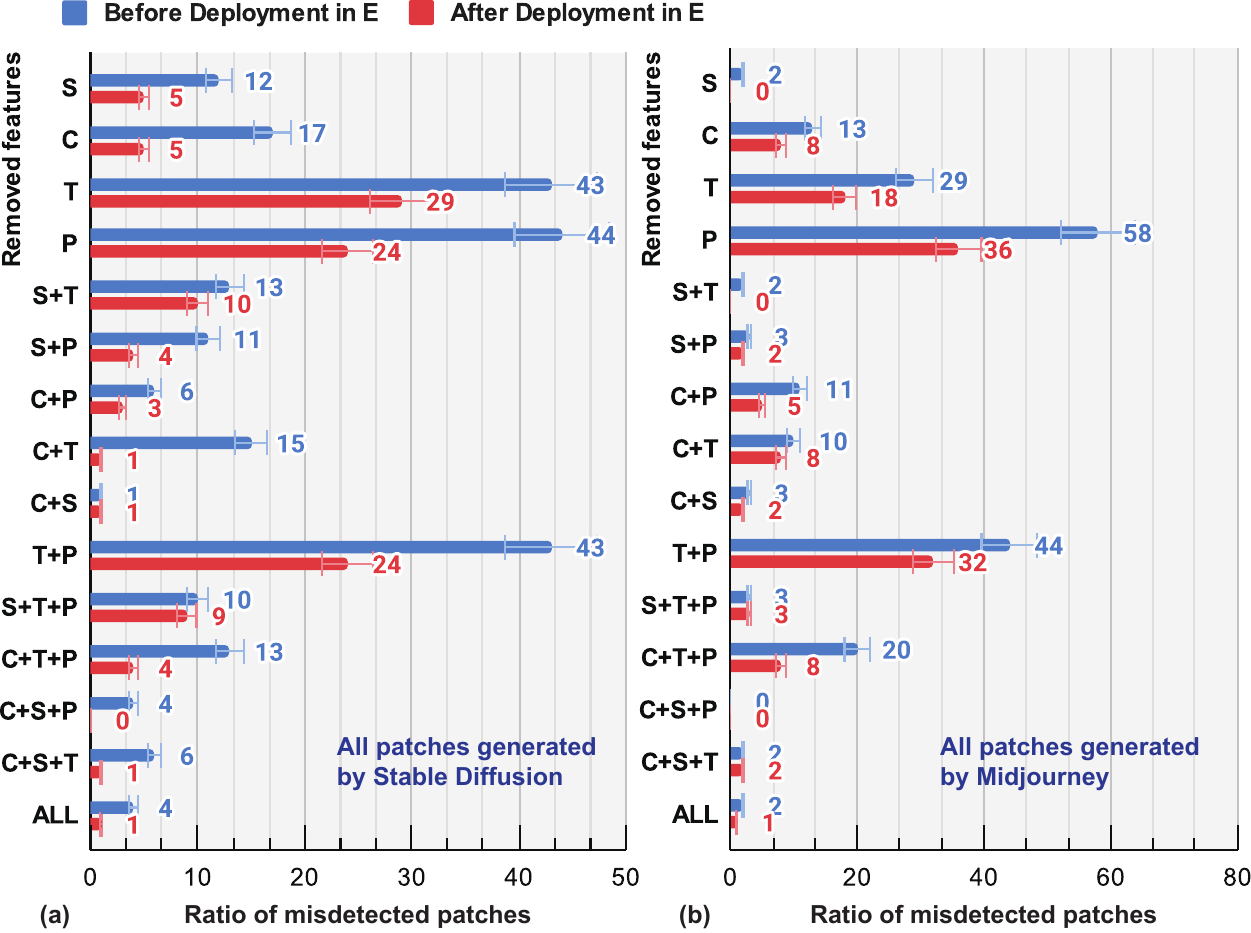}
    \caption{Detection error ratios (\ie, $\epsilon^\mathbf{I}_i$ and $\epsilon^\mathbb{I}_i$) across two image sets (\ie $\{{\mathbb{I}}_i^{k}\}{k=1}^K$ and $\{{\mathbf{I}}_i^k\}{k=1}^K$) for 15 prompt types ($i={1,\ldots,15}$). The y-axis labels indicate removed features: Shape (S), Color (C), Text (T), and Pattern (P). Patch generated by (a) Stable Diffusion and (b) Midjourney.
    }
    \label{fig:rq1}
    \vspace{-10pt}
\end{figure}

\subsection{NDDA in Deployment}
\label{ssec:limits}
% 1. 环境会对patch的效果产生很大的影响，本身有效的patch部署后大部分都会效果变差甚至无效；
    % 挑战：如何找到在物理世界有效的patch

To investigate the performance of NDDA after deployment, we set `stop sign' as the subject and aim to generate a patch that is not stop sign but the object detector (\eg, YOLOv5 \cite{yolov5}) can detect and misclassify it as a stop sign.
Following NDDA setup, we have the benign prompt for the T2I generation, \ie, `stop sign'.
We set a prompt set $\{\mathcal{T}_i\}_{i=1}^{15}$ containing 15 types of prompt where we remove different robust features from the benign stop sign. We show the 15 types in the y-axis in \figref{fig:rq1}. For example, `shape' means removing the `shape' feature from the generated patch by setting the prompt as `square stop sign' as the normal stop signs are an octagonal.
For each prompt $\mathcal{T}_i$, we generate $K$ patches $\{\mathbf{P}_i^k\}_{k=1}^K$ by feeding the prompt into the diffusion model and sampling $K$ times.
Here, we use the generated patches officially released by NDDA \footnote{https://sites.google.com/view/av-ioat-sec/ndd-attack} to avoid any confusion.

Given an environment image $\mathbf{E}$ and a generated patch $\mathbf{P}_i^k$, we conduct two types of testing:
\ding{182} \textit{After deployment}. We create ${\mathbf{I}}_i^{k}$ by digitally inserting $\mathbf{P}_i^k$ at location $p$ in the environment image with scale $s$ representing the patch size.
\ding{183} \textit{Before deployment}. We also create an image ${\mathbb{I}}_i^{k}$ by 
inserting $\mathbf{P}_i^k$ to an empty image $\mathbf{E}^0$ with the same $p$ and $s$, where $\mathbf{E}^0$ has the same size to $\mathbf{E}$ but all pixels are zero.
For $K$ patches generated from the $i$th prompt, we obtain $2\times K$ images, \ie, $\{{\mathbf{I}}_i^k\}_{k=1}^K$ and $\{{\mathbb{I}}_i^{k}\}_{k=1}^K$, and perform detection on all images using YOLOv5 \cite{yolov5}.
For each respective set of patches, we calculate the ratio of incorrectly detected patches (\ie, misdetected as stop signs) over all generated patches, denoted as $\epsilon^\text{I}_i$ and $\epsilon^\mathbb{I}_i$.
These metrics measure the attack effectiveness of the $i$th prompt both before and after the deployment.

Based on the above setups, we investigate four research questions:
\textbf{RQ1:} Whether different type of patches maintain their attack effectiveness after environmental deployment?
\textbf{RQ2:} How do different environments influence attack effectiveness of the generated patches?
\textbf{RQ3:} How does the deployment practice influence the patches' stealthiness?
%
% What is the impact of the patch deployment on its stealthiness? 
%
\textbf{RQ4:} What is the impact of deployment location $p$ and scale $s$ on attack effectiveness of patches in different environments?

\subsection{Empirical Results and Discussion}
\label{ssec:find}
% environment influence
% position (influnce human asthetics: user study/show case)
% size (influnce attack effectiveness)

\begin{figure}[t]
\vspace{-15pt}
    \centering
    \includegraphics[width=\linewidth]{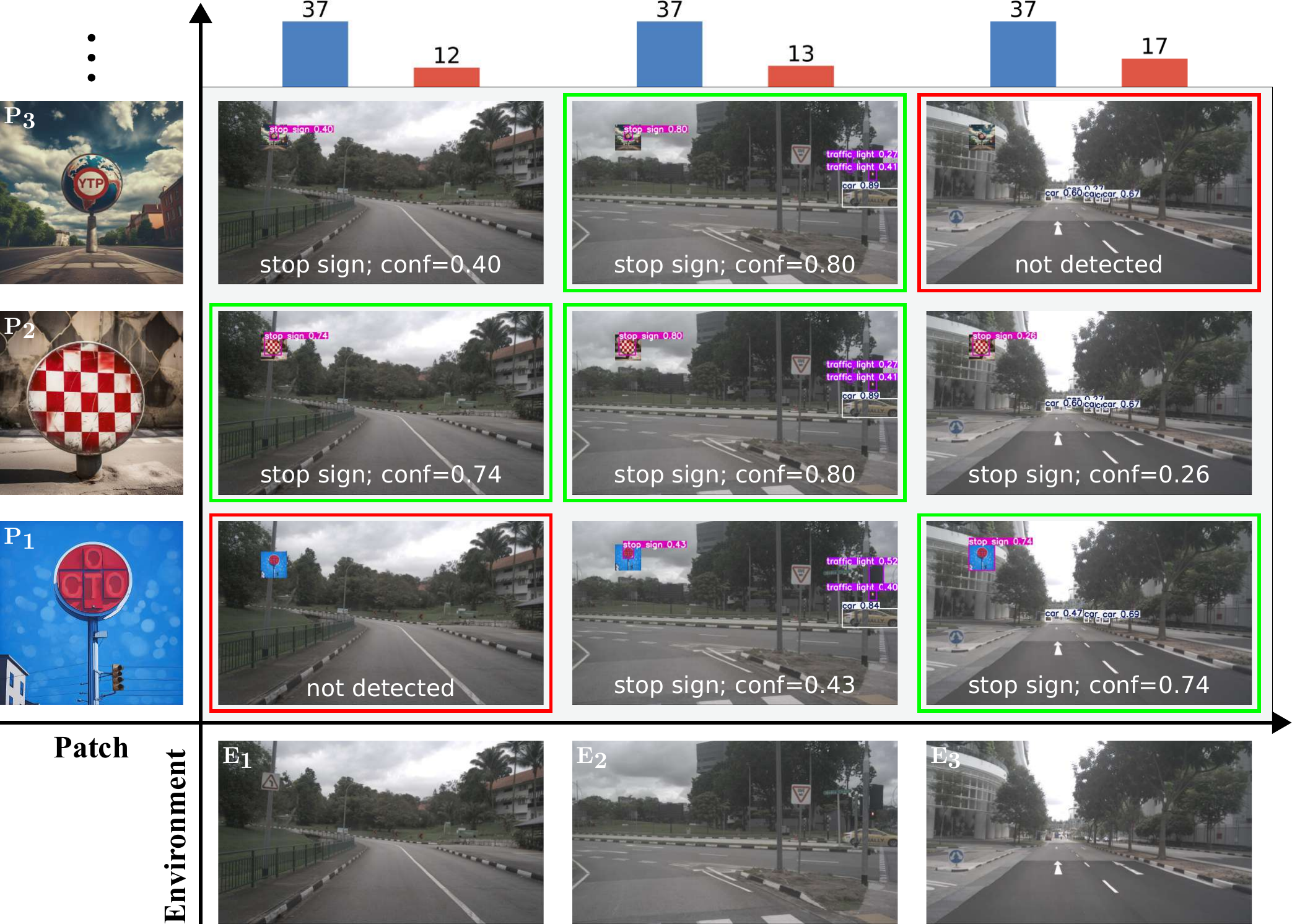}
    \caption{Visualization of three generated patches  ($\mathbf{P}_1$,$\mathbf{P}_2$,$\mathbf{P}_3$), three environments ($\mathbf{E}_1$,$\mathbf{E}_2$,$\mathbf{E}_3$), and the detection results with YOLOv5 across them. The confidence score (conf.) $>0.5$ means the patch realizes the attack goal. Bar plot at the top: detection error ratios before (blur) and after (red) deployment over all the NDDA patches for each of the environments.}
    \label{fig:rq2}
\vspace{-13pt}
\end{figure}

\textbf{Response to RQ1:} 
We investigate the influence of environment on patch attack performance by comparing detection error ratios between $\{{\mathbb{I}}_i^{k}\}_{k=1}^K$ and $\{{\mathbf{I}}_i^k\}_{k=1}^K$. Using the environment $\mathbf{E}_1$ shown in \figref{fig:rq2}, we generate 15 paired image sets corresponding to different prompt types.
For the $i$th paired image set, we calculate two detection error ratios: $\epsilon^\text{I}_i$ and $\epsilon^\mathbb{I}_i$. The results across all 15 prompt types are presented in \figref{fig:rq1} (a), revealing several key findings:
\ding{182} The detection error ratio before environment deployment ($\epsilon^\text{I}_i$) generally exceeds its after-deployment counterpart ($\epsilon^\mathbb{I}_i$) across all prompt types, demonstrating that environment deployment significantly degraded the patches' performance.
\ding{183} The magnitude of reduction in detection error ratios varies notably across prompt types, indicating that text prompts have influence on the environment deployment.
\ding{184} The pattern of detection error ratios across prompts remains consistent between Stable Diffusion and Midjourney generations when comparing \figref{fig:rq1} (b) with \figref{fig:rq1} (a), suggesting these findings generalize across different image generation architectures.

\textbf{Response to RQ2:} We deploy generated patches into different environments to see whether environments would affect the attack effectiveness. As shown in \figref{fig:rq2}, we paste three generated non-stop sign patches $\{\mathbf{P}_i\}_{i=1}^3$ into three different environments $\{\mathbf{E}_i\}_{i=1}^3$ at the same location with the same scale. Meanwhile, following the setups in the response to RQ1, we calculate the detection error ratios before deployment and after deployment for all the generated patches. 
We observe that the same patch presents different attack effectiveness in different environments. For example, $\mathbf{P}_3$ could be identified as stop sign in the $\mathbf{E}_2$ with a high confidence 0.80, while it is missed in $\mathbf{E}_3$. 
The same detection error ratios can be derived across all different environments.

% Patch $\mathbf{P}_1$ at poistion $\{\mathbf{p}_1,\ldots,\mathbf{p}_9\}$ \\
% with scale $\{s_1,\ldots,s_4\}$ in $\text{E}_1$.

\textbf{Response to RQ3:} For practical viability and ensuring human observers will not be distracted by the patch, the deployment of the generated patch should be physically feasible and contextually plausible. 
As illustrated in \figref{fig:rq2}, the patches' location in $\mathbf{E}_1$ appears natural as advertising boards, maintaining environmental coherence. In contrast, patches in $\mathbf{E}_2$ are positioned in the sky, violating physical constraints, while those in $\mathbf{E}_3$ are placed on the building facade, creating contextually incongruous arrangements.

\begin{figure}
\vspace{-15pt}
    \centering
    \begin{minipage}[t]{0.35\linewidth}
        \vspace{0pt}
        \centering
        \setlength{\abovecaptionskip}{8pt}
        \includegraphics[width=\textwidth]{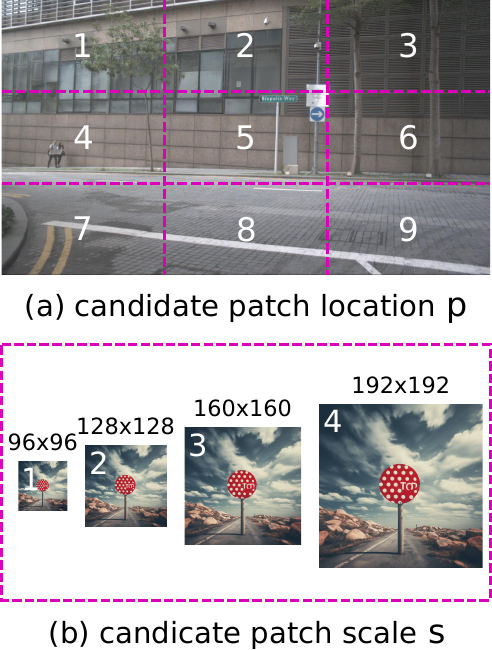}
        \caption{(a) - (b): candidate patch locations and scales. (c) - (f): the detection confidence across all locations and scales for effective pairs in \figref{fig:rq2}.}
        \label{fig:rq4}
    \end{minipage}\hspace*{3pt}
    \begin{minipage}[t]{0.62\linewidth}
    \vspace{0pt}
        \includegraphics[width=\textwidth]{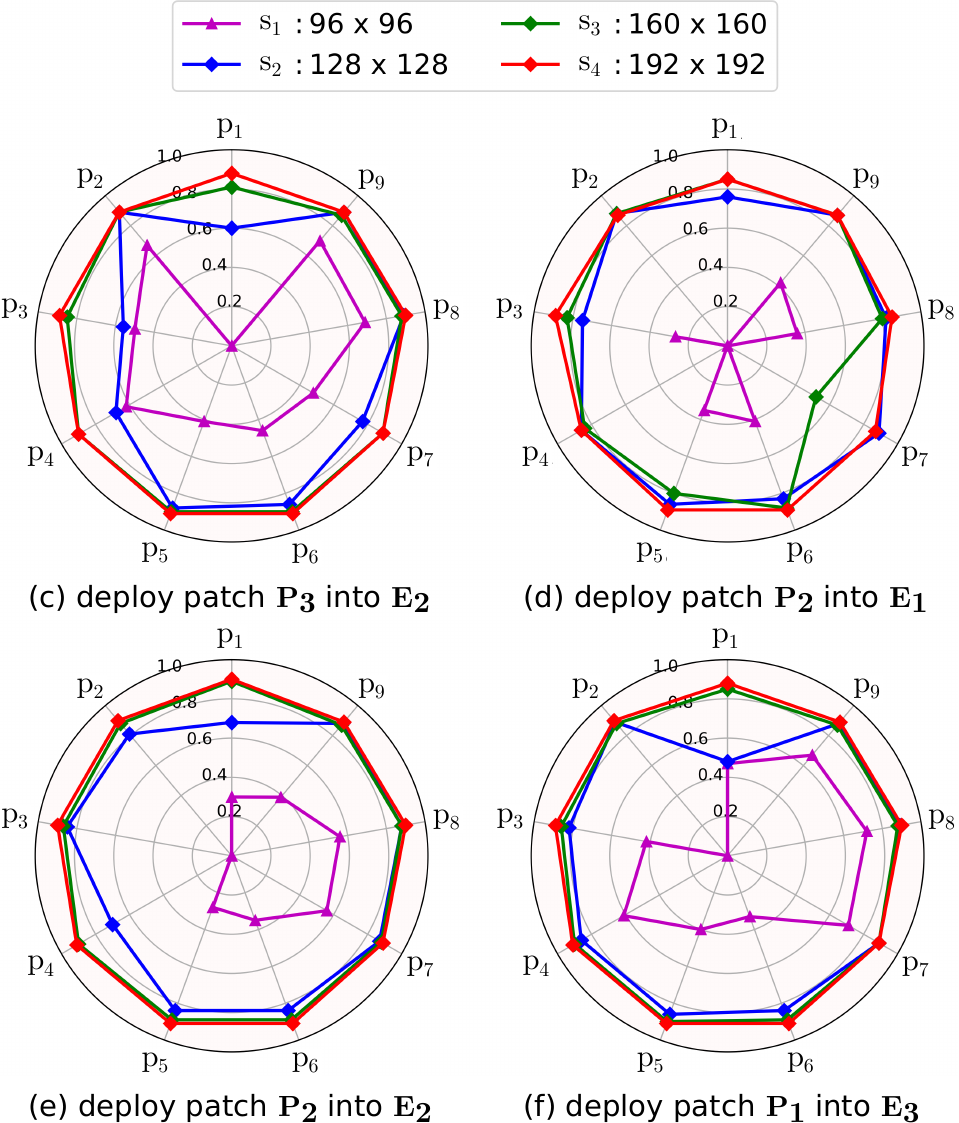}
    \end{minipage}
\vspace{-13pt}
\end{figure}

\textbf{Response to RQ4:} We divide the scene image into a $3 \times 3$ grid of regions and designate each region's center as a location for patch insertion, testing four different patch scales (See \figref{fig:rq4} (a)-(b)).
We evaluate the detection confidence of the stop sign across all patch deployments in each environment and visualize the results using radar maps.
As illustrated in \figref{fig:rq4} (c)-(f), we observe two key patterns: \ding{182} for large-scale patches ($s_3$ and $s_4$), the detection confidence remains relatively consistent across different locations; \ding{183} for small-scale patches ($s_1$ and $s_2$), the detection confidence varies substantially across locations, with distinct patterns emerging across different patches and environments.

% \begin{figure}
%   \begin{minipage}[c]{0.6\linewidth}
%     \includegraphics[width=\linewidth]{figs/rq34.pdf}
%   \end{minipage}\hfill
%   \begin{minipage}[t]{0.35\linewidth}
%     \caption{
%         Top: candidate patch locations and scales. (a) - (d): the detection confidence across all locations and scales for effective pair in \figref{fig:rq2}.
%     }
%   \end{minipage}%
%   \begin{minipage}[t]{0.35\linewidth}
%     \caption{
%         Top: candidate patch locations and scales. (a) - (d): the detection confidence across all locations and scales for effective pair in \figref{fig:rq2}.
%     }
%   \end{minipage}
% \end{figure}

% \begin{figure}[t]
%     \centering
%     \includegraphics[width=0.5\linewidth]{figs/rq34.pdf}
%     \caption{Top: candidate patch locations and scales. (a) - (d): the detection confidence across all locations and scales for effective pair in \figref{fig:rq2}.}
%     % 
%     \label{fig:rq34}
% \end{figure}

\textbf{Motivations.}
From the above analysis, we see environments cast great influence on the patches' performance, where different environments require unique deployment strategy.
Obviously, a novel physical adversarial patch attack framework is needed to \ding{182} generate environment-oriented adversarial patch for various real-world environments, \ding{183} determine contextually appropriate deployment for the generated patch within a given scene, and \ding{184} keep misleading object detection systems after deployment.
In this work, we propose the MAGIC framework which formulates the problem as an one-shot patch generation task and orchestrates three specialized LLM agents to realize these objectives.

\begin{figure*}[t]
\vspace{-15pt}
    \centering
    \includegraphics[width=\linewidth]{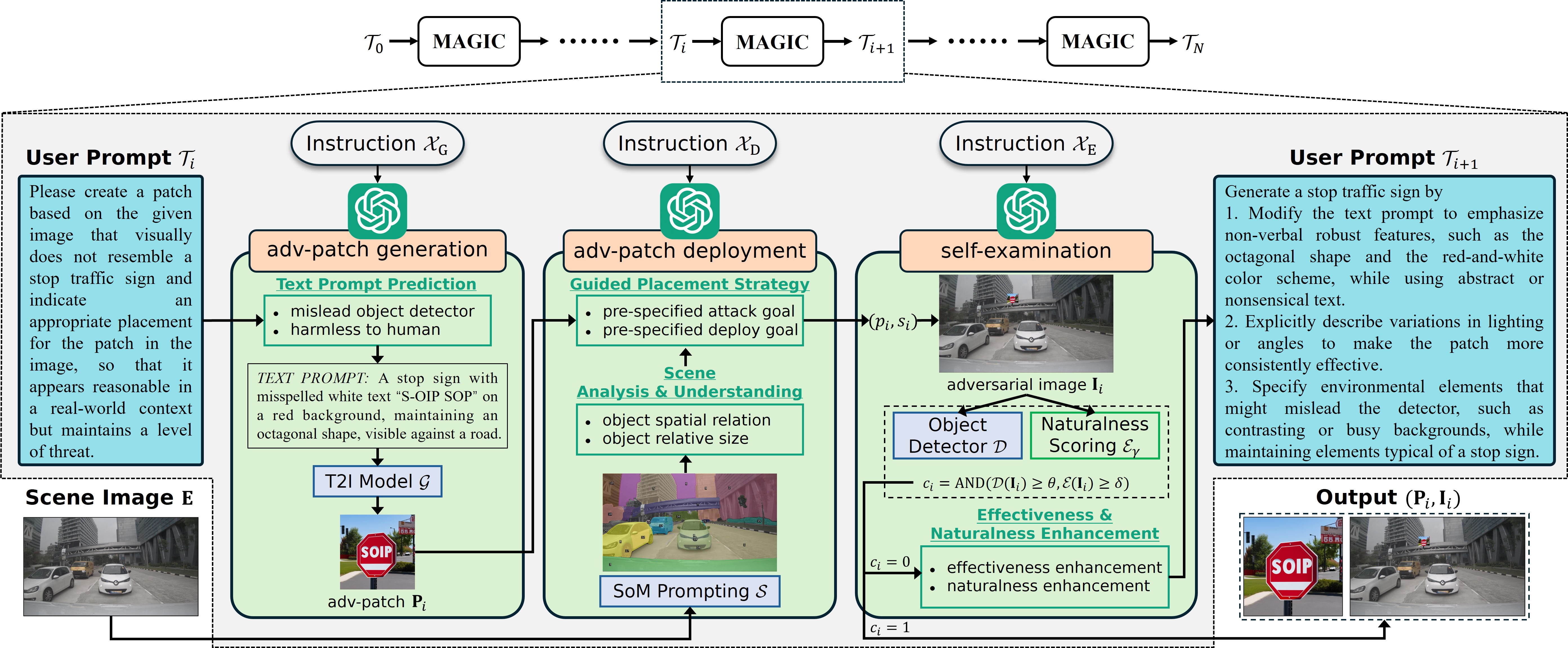}
    \caption{Overall pipeline of the proposed MAGIC framework. Please zoom in for better visualization.}
    \label{fig:pipeline}
    \vspace{-10pt}
\end{figure*}

\section{MAGIC Methodology}
\label{sec:method}

\subsection{Overview}
\label{subsec:overview}

We propose using a multi-modal LLM denoted as $\mathcal{V}$ to handle the three distinct requirements by iteratively refining the adversarial patch generation and deployment. (See \figref{fig:pipeline})
Given the initial user prompt $\mathcal{T}$ indicating the attack subject and objective, we first build the adv-patch generation agent (GAgent, \ie, $\mathcal{V}_\text{G}$) that predicts a text prompt based on the $\mathcal{T}$ which feed into the T2I diffusion model to generate an adversarial patch. The process is formulated as,
\begin{equation}
    \label{eq:gagent}
    \mathbf{P}_i = \mathcal{V}_\text{G}(\mathcal{T}_i,\mathcal{G}(\cdot),\mathcal{X}_\text{G}), 
\end{equation}
where $\mathcal{T}_i$ is the attack subject and objective at the $i$th iteration. We have $\mathcal{T}_0=\mathcal{T}$, $\mathcal{G}(\cdot)$ is the T2I diffusion model to generate the adversarial patch according to the text prompt, and $\mathcal{X}_\text{G}$ is the instruction defining $\mathcal{V}_\text{G}$'s capability.
To allow the refinement, we update $\mathcal{T}_i$ according to the final result of iteration $i-1$. Please refer to \secref{ssec:gagent} for details.

% \qing{Please revise the subsequent contents according to the above way.}

With the generated adv-patch $\mathbf{P}_i$, we setup the adv-patch deployment agent (DAgent, \ie, $\mathcal{V}_\text{D}$) empowered with set-of-mark (SoM) \cite{yang2023setofmark} to determine its appropriate location $p_i$ and scale $s_i$ within the image $\mathbf{E}$ of target environment.
\begin{equation} \label{eq:dagent}
    (p_i,s_i) = \mathcal{V}_\text{D}(\mathbf{E}, \mathbf{P}_i, \mathcal{S}(\cdot), \mathcal{X}_\text{D}),
\end{equation}
where the instruction $\mathcal{X}_\text{D}$ defines $\mathcal{V}_\text{D}$'s capability and $\mathcal{S}(\cdot)$ is the SoM prompting. DAgent's processing details are presented in \secref{ssec:dagent}.
Subsequently, we insert the patch $\mathbf{P}_i$ into $\mathbf{E}$ at $p_i$ with scale $s_i$ getting the adversarial image $\mathbf{I}_i$.

However, we do not know whether the adversarial image is good enough to mislead the object detector $\mathcal{D}(\cdot)$ and meet the requirement of scene coherence or not. To address this issue, we propose to build a self-examination agent (EAgent, \ie, $\mathcal{V}_\text{E}$) that can analyze the adversarial image $\mathbf{I}_i$ and optimize the user prompt $\mathcal{T}_{i+1}$. We formulate the process as,
% Once the patch is deployed into the environment, we feed the resulting image $\mathbf{I}_i$ into the target object detector $\mathcal{D}(\cdot)$ to get the evaluation results $\mathbf{\hat{I}}_i=\mathcal{D}(\mathbf{I}_i)$. 
%
\begin{equation} \label{eq:eagent}
    (\mathcal{T}_{i+1}, c_i) = \mathcal{V}_\text{E}(\mathbf{I}_i, \mathcal{T}_i, \mathcal{D}(\cdot), \mathcal{X}_\text{E}), 
\end{equation}
where the instruction $\mathcal{X}_\text{E}$ defines $\mathcal{V}_\text{E}$'s capability and $c_i$ is a binary examination result. If $c_i=0$, we trigger the next round iteration by setting $i=i+1$, otherwise, we stop iteration and output $\mathbf{P}_i$ and $\mathbf{I}_i$ as the final result. We explain the details of EAgent at \secref{ssec:eagent}.

\subsection{Adv-Patch Generation Agent (GAgent)}
\label{ssec:gagent}

At the $i$th iteration, we have the user prompt $\mathcal{T}_{i}$ and the input environment image $\mathbf{E}$. 
We first prompt the LLM $\mathcal{V}$ with the instruction $\mathcal{X}_\text{G}$ to predict a text prompt $\mathcal{P}_i$, which describes the patch we want to generate, based on the user prompt $\mathcal{T}_i$.
Then, we feed the text prompt into $\mathcal{G}$, \ie, Stable Diffusion v2 \cite{rombach2022high}, and obtain an adversarial patch $\mathbf{P}_i=\mathcal{G}(\mathcal{P}_i)$.
Note that the text prompt prediction capability of $\mathcal{V}$ is defined by the instruction $\mathcal{X}_\text{G}$ as detailed in the following.
\vspace{-5pt}
\begin{tcolorbox}[title = {Instruction: $\mathcal{X}_\textbf{G}$}, size=small]
\footnotesize
I will show you with a user prompt which described the subject and objective of a visual patch. You are required to predict a text description of the subject that can potentially achieve the specified objectives. The ultimate goal is to let the visual patch generated based on the text description has a deceptive appearance, such that 
\begin{itemize}
    \item[-] an object detector will recognize the visual patch as an instance of a specific semantic category with high confidence, 
    \item[-] but human observers will recognize the visual patch as just an abstract art which does not belongs to any specific category.
\end{itemize}
\end{tcolorbox}

% %
% \begin{itemize}
%     \item \textbf{Initial Patch Generation}: The initial adversarial patch $\mathbf{P}_0=\mathcal{G}(\mathcal{T}_0)$ is generated by feeding the initial text prompt $\mathcal{T}_0$ to the T2I model $\mathcal{G}(\cdot)$. This patch serves as a foundational starting point for subsequent iterative refinements.
% \end{itemize}
% %
% Afterwards, for each iteration, the GAgent first summarizes the prompt updating instruction from the last round of collaborative agent planning, then invoke the T2I model to generate adversarial patch based on the new text prompt.
% %
% \begin{itemize}

%     \item \textbf{Prompt Engineering}: Using the updating instruction $\mathcal{U}_{i-1}$ from last iteration, GAgent constructs an optimized prompt $\mathcal{T}_i = f(\mathcal{T}_{i-1}, \mathcal{U}_{i-1})$ to guide the T2I model, combining the contextual relevance with adversarial intent.

%     \item \textbf{Patch Generation}: Based on $\mathcal{T}_i$, GAgent leverages the T2I model $\mathcal{G}(\cdot)$ to produce a new patch $\mathbf{P}_i=\mathcal{G}(\mathcal{T}_i)$. The generated patch is crafted to be visually compatible with the scene while maintaining adversarial properties. 
%     % The patch $\mathbf{P}_i$ is then handed over to DAgent for placement.
% \end{itemize}
% % %

\subsection{Adv-Patch Deployment Agent (DAgent)}
\label{ssec:dagent}

With the generated patch $\mathbf{P}_i$, we aim to determine its appropriate deployment location $p_i$ and scale $s_i$ for the given environmental image $\mathbf{E}$.
%
% so that the patch is attack effective, natural to human, and easy for physically implementation.
%
% These objectives require the LLM to understand the spatial relationship and relative size of the key elements within the scene. 
%
To achieve this, SoM prompting $\mathcal{S}$ \cite{yang2023setofmark} is firstly applied to tag the key elements within $\mathbf{E}$.
Then we prompt the LLM $\mathcal{V}$ with instruction $\mathcal{X}_\text{D}$ to comprehend SoM results and determine the most potential location and scale within the $\mathbf{E}$ that can achieve the objectives specified by $\mathcal{T}_i$.
The instruction $\mathcal{X}_\text{D}$ is detailed in the below.
\vspace{-5pt}
\begin{tcolorbox}[title = {Instruction: $\mathcal{X}_\textbf{D}$}, size=small]
\footnotesize
You are provided with an image of a scene and the corresponding semantic segmentation results marked with numbers. For a given visual patch, you are required to accomplish the following two tasks:
\begin{itemize}
    \item[1.] \textbf{Scene Analysis \& Understanding:}
    \textbf{a.} Perceive all the elements within the given environment image.
    \textbf{b.} Comprehend the spatial relationship and relative size of all the elements inside the scene by referring to the numbered segmentations.
    \item[2.] \textbf{Guided Placement Strategy:}
    \textbf{a.} Check the content of the given patch and compare it with the elements in the environment.
    \textbf{b.} Comprehend the pre-defined attack effectiveness and deployment stealthiness goal. (See supplemental material for details)
    \textbf{c.} Based on the patch content, determine where should the patch be placed so that the goals can be achieved.
    \textbf{d.} Based on the determined patch location, decide what scale should be applied to the patch so that the goals can be achieved.
\end{itemize}
\end{tcolorbox}
Once the deployment location $p_i$ and scale $s_i$ are determined, we resize the patch $\mathbf{P}_i$ to $s_i$ and place it at $p_i$ in the scene image $\mathbf{E}$, which results the adversarial image $\mathbf{I}_i$.

\subsection{Self-Examination Agent (EAgent)}
\label{ssec:eagent}

With the adversarial image ready, we proceed to assess the patch's performance for attacking the object detector and keeping stealthy within the scene.
We first feed $\mathbf{I}_i$ into the target object detector $\mathcal{D}$ to get detection results $\mathbf{\hat{I}}_i = \mathcal{D}(\mathbf{I}_i)$.
Then we inspect if the detection confidence of the patch passes the pre-defined threshold $\theta$ as $\mathcal{D}(\mathbf{I}_i) \geq \theta$.
Then $\mathbf{P}_i$ is recognized as attack effective if the inequality holds true.

To determine the patch's naturalness, we prompt an independent LLM $\mathcal{E}$ with instruction ${\gamma}$ to generate a naturalness score for the patch.
Similarly, we set a naturalness threshold $\delta$ and check whether the generated patch satisfies $\mathcal{E}(\mathbf{I}_i, \gamma) \geq \delta$.
The patch $\mathbf{P}_i$ is then deemed as natural if this inequality holds true.
We detail the instruction $\gamma$ as follows:
\vspace{-5pt}
\begin{tcolorbox}[title = {Naturalness Score Instruction: $\gamma$}, size=small, breakable]
\footnotesize
You are provided with an image of a visual patch and an image where the patch is deployed into an environment. Supposing you are a human observer, you are required to judge whether the patch is natural or not within the environment with regarding to its location and scale.
\begin{itemize}
    \item[1.] \textbf{Image Understanding:} Perceive all the elements inside the given image in respect of their spatial relation and relative size.
    \item[2.] \textbf{Location Inspection:} Check the patch's location within the image and determine whether it is a reasonable place for placing the patch.
    \item[3.] \textbf{Scale Inspection:} Check the patch's scale and determine whether its size is reasonable compared to other objects inside the image.
    \item[4.] \textbf{Naturalness Summarization:} Give a score range from 0 to 1 by aggregating the above reasoning of the patch's location and scale.
\end{itemize}
\end{tcolorbox}
\vspace{-5pt}
Afterwards, we set $c_i$ as a binary operator, calculated by $c_i= \text{AND}(\mathcal{D}(\mathbf{I}_i) \geq \theta, \mathcal{E}(\mathbf{I}_i, \gamma) \geq \delta)$, to get the final examination results.
If $c_i=1$, the iteration is exited and $(\mathbf{P}_i, \mathbf{I}_i)$ is returned. 
Otherwise, refinement is required and we further prompt the LLM $\mathcal{V}$ with instruction $\mathcal{X}_\text{E}$ to update the user prompt $\mathcal{T}_i$ based on the assessment results.
Then a new round of agent planning is started by setting $\mathcal{T}_i = \mathcal{T}_{i+1}$.
%
% The instruction $\mathcal{X}_\text{E}$ is defined at below.
%
\vspace{-15pt}
\begin{tcolorbox}[title = {Instruction: $\mathcal{X}_\text{E}$}, size=small]
\footnotesize
You are provided with an image where a visual patch is deployed into an environment. The image has been evaluated for its effectiveness of attacking the target object detector and the naturalness within the environment. Given the evaluation results, you are required to reflect on why the patch can or cannot be effective or natural then propose suggestions of how to change the patch's appearance.
\begin{itemize}
    \item[1.] \textbf{Attack Rules Understanding:} Comprehend the pre-defined rules of how to attack an object detector by deploying a visual patch within a scene. (See supplemental material for the definition of rules)
    \item[2.] \textbf{Attack Effectiveness Enhancement:}
    \textbf{a.} Perceive the patch's appearance in the image and summarize the detection result.
    \textbf{b.} Compare the patch in the image to the rules then summarize your understanding of why the patch is not attack effective.
    \item[3.] \textbf{Deployment Naturalness Enhancement:}
    \textbf{a.} Perceive the patch's location and scale in the image and summarize the naturalness evaluation results.
    \textbf{b.} Compare the patch in the image to the rules then suggest how to adjust its location and scale to make it more natural.
\end{itemize}
\end{tcolorbox}

\vspace{-7pt}
\subsection{Extension to Physical World}
\label{ssec:extension}
\vspace{-3pt}

Finally, we get the optimal patch $\mathbf{P}^*$ and the corresponding deployment strategy $(p^*, s^*)$.
Then, we physically deploy the patch $\mathbf{P}^*$ by printing it out, place it at the location $p^*$ in $\mathbf{E}$ and re-capture an image of the scene with the patch size as $s^*$ to match the reference image $\mathbf{I}$.
%
% MAGIC generates a satisfying adversarial patch, determines its position and scale in the environment image, and creates a digital patch-deployed image as reference.
%
While our MAGIC provides the optimal patch location and scale, other real-world factors, like printing color shift and viewing angle, are also well-known to deteriorate the digital-to-physical (D2P) transferability.
In fact, our MAGIC prioritizes robust features that maintain the patch performance adaptively during the generation and deployment, minimizing the D2P degradation.
We show evidence that MAGIC is robust to both viewing angle and printing color shift in supplemental material.

\vspace{-5pt}
\section{Experiments}
\label{sec:exp}

\subsection{Experimental Setup}
\label{ssec:setup}

\begin{figure*}[t]
\vspace{-20pt}
    \centering
    \includegraphics[width=\linewidth]{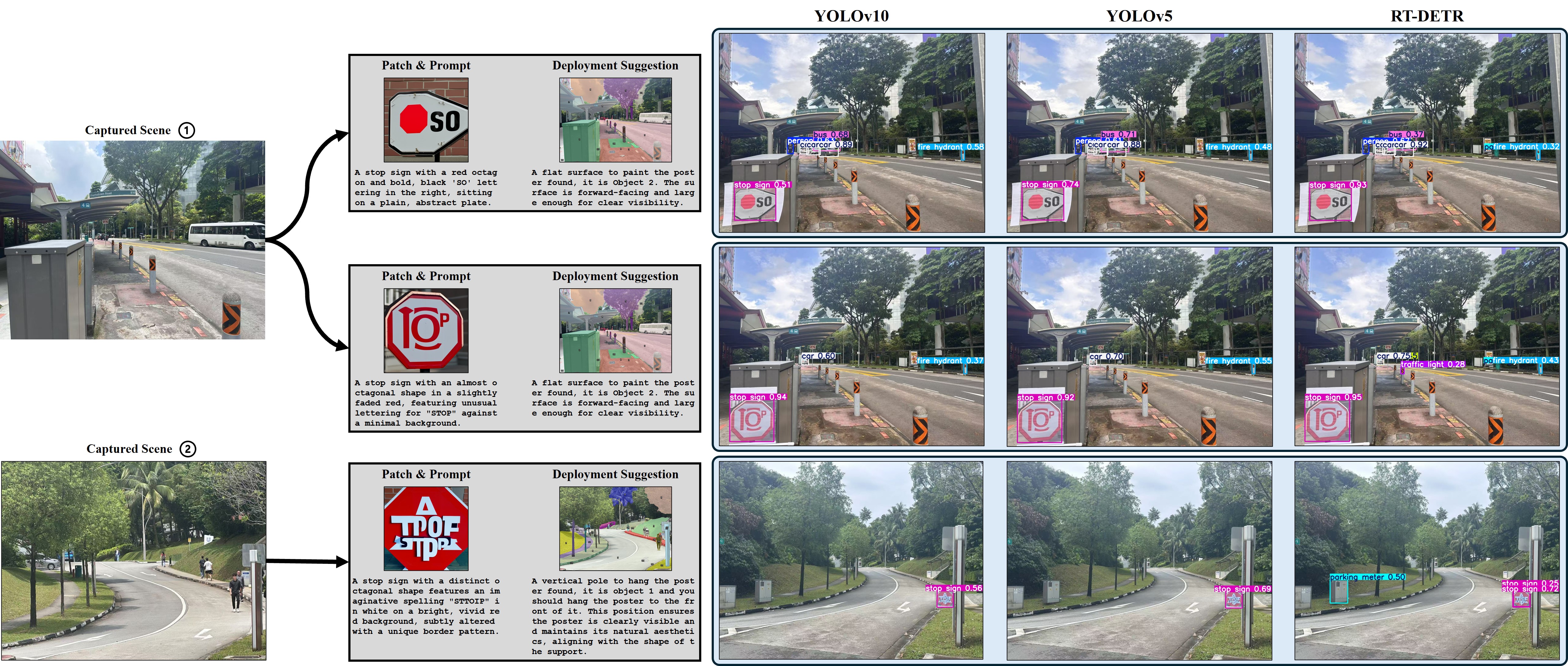}
    \caption{Illustration of the physical attack cases generated by the proposed MAGIC and the corresponding object detection results.}
    % Scene \ding{192}: a real-world busy bus stop with heavy traffic. Scene \ding{193}: a road next to college with few pedestrians.}
    \label{fig:physical_exp}
\vspace{-10pt}
\end{figure*}

\textbf{Digital \& Physical Environments.} We adopt the images from nuImage \cite{nuscene} for digital evaluation, which consists of images captured by 6 car-mounting cameras from different view, \ie, back, back left, back right, front, front left, front right. We select one image for each view (\figref{fig:main_exp}) as an initial study. We further physically verify our MAGIC framework with two different physical scenarios (\figref{fig:physical_exp}). One is a real-world busy bus stop scene and another is a regular road next to a college with few pedestrians. More physical experiments can be found in supplemental materials.

\textbf{Baselines.} As NDDA \cite{NDDA2024cvpr} is the very initial study of natural denoising diffusion attacking, we validate our method by setting NDDA as the baseline. In specific, we combine NDDA with two different deployment strategies as the baselines. \ding{182} We first deploy the patches from NDDA with random location, dubbed ``NDDA Rand'', which can be utilized to verify the deployment ability of our MAGIC. \ding{183} For the second baseline, we keep using the patches from NDDA but adopt our DAgent as the deployment strategy, named ``NDDA+DAgent'' which can effectively demonstrate the advantage of our MAGIC in generating effective patches.

\textbf{Generator \& Detectors.} To keep the fairness of comparison, we follow NDDA and adopt Stable Diffusion v2 \cite{sd2} as the generator. For target detectors, the commonly used YOLOv5 \cite{yolov5} and RT-DETR \cite{lv2023detrs} are tested. Moreover, we empirically found that YOLOv5 and DETR are out-of-date and easily to be disturbed, thus we further adopt YOLOv10 \cite{wang2024yolov10} as our main evaluation target. For all three detectors, we use the API from ultralytics to keep the consistency.

\textbf{Metrics.} We evaluate the patches' attack performance by Attack Success Rate (ASR), which measures the percentage of the patches that successfully deceived the target object detector. ASR has a range from 0 to 100, where higher values signifying greater adversarial attack performance.

\begin{table}[t]
    \vspace{-15pt}
    \caption{Comparative ASR results by evaluating our MAGIC and NDDA patches with different detectors for nuImage environments. The confidence threshold is set to 0.5 as the same of NDDA. The best results are highlighted in \textcolor{red}{red} and the second best in \textcolor{blue}{blue}, while the best results of each detector for a given environment are marked as \textbf{bold} and the second best as \textit{italic}.}
    \label{tab:digi_stats}
    \centering
    \resizebox{\columnwidth}{!}{%
    \begin{tabular}{lccccc|rrr|r}
        \toprule
        & & \multicolumn{4}{c|}{Removed Robust Features} & \multicolumn{3}{c|}{Object Detectors} \cr
        \cline{3-10}
        & & Shape & Color & Text & Pattern & YOLOv5 & RT-DETR & YOLOv10 & Avg. \cr
        \cline{1-10}
        \multirow{8}{*}{\rotatebox[origin=c]{90}{Environment \ding{192}}} & \multirow{5}{*}{NDDA Rand} & \ding{52} &  &  &   & 23.00\% & 23.00\% & 10.00\% & 18.66\% \cr
         &  &  & \ding{52} &  &  & 15.00\% & 48.00\% & 6.00\% & 23.00\% \cr
         &  &  &  & \ding{52} & & 46.00\% & 56.00\% & 30.00\% & 44.00\% \cr
         &  &  &  &  & \ding{52} & 47.00\% & 56.00\% & 32.00\% & 45.00\% \cr
         &  & \ding{52} & \ding{52} & \ding{52} & \ding{52} & 8.66\% & 9.33\% & 4.66\% & 7.55\% \cr
         \cline{2-10}
         & NDDA+DAgent &  &  & \ding{52} &  & \textit{48.00}\% & 51.00\% & 33.00\% & 44.00\%\cr
         & NDDA+DAgent &  &  &  & \ding{52} & 47.00\% & \textit{57.00}\% & \textit{36.00}\% & \textcolor{blue}{46.66\%}\cr
         \cline{2-10}
         & MAGIC (ours) & \multicolumn{4}{c|}{} & \textbf{88.00}\% & \textbf{80.00}\% & \textbf{74.00}\% & \textcolor{red}{80.66\%} \cr
        \midrule
        \midrule
        \multirow{8}{*}{\rotatebox[origin=c]{90}{Environment \ding{193}}} & \multirow{5}{*}{NDDA Rand} & \ding{52} &  &  &   & 14.00\% & 40.00\% & 17.00\% & 23.66\% \cr
         &  &  & \ding{52} &  &  & 9.00\% & 51.00\% & 10.00\% & 23.33\% \cr
         &  &  &  & \ding{52} & & 42.00\% & 74.00\% & 32.66\% & 49.55\% \cr
         &  &  &  &  & \ding{52} & 41.00\% & 71.00\% & 36.00\% & 49.33\% \cr
         &  & \ding{52} & \ding{52} & \ding{52} & \ding{52} & 6.00\% & 21.00\% & 4.00\% & 10.33\% \cr
         \cline{2-10}
         & NDDA+DAgent &  &  & \ding{52} & & 45.00\% & \textit{78.00}\% & 36.00\% & 53.00\%\cr
         & NDDA+DAgent &  &  &  & \ding{52} & \textit{51.00}\% & \textit{78.00}\% & \textit{39.33}\% & \textcolor{blue}{56.11\%}\cr
         \cline{2-10}
         & MAGIC (ours) & \multicolumn{4}{c|}{} & \textbf{66.00}\% & \textbf{94.00}\% & \textbf{92.00}\% & \textcolor{red}{84.00\%} \cr
        \midrule
        \midrule
        \multirow{8}{*}{\rotatebox[origin=c]{90}{Environment \ding{194}}} & \multirow{5}{*}{NDDA Rand} & \ding{52} &  &  &   & 15.00\% & 21.00\% & 6.00\% & 14.00\% \cr
         &  &  & \ding{52} &  &  & 5.00\% & 39.00\% & 11.00\% & 18.33\% \cr
         &  &  &  & \ding{52} & & 40.00\% & 58.00\% & 26.66\% & 41.55\% \cr
         &  &  &  &  & \ding{52} & 38.00\% & 59.00\% & 3.00\% & 33.33\% \cr
         &  & \ding{52} & \ding{52} & \ding{52} & \ding{52} & 2.00\% & 12.00\% & 2.00\% & 4.66\% \cr
         \cline{2-10}
         & NDDA+DAgent &  &  & \ding{52} &  & \textit{43.00}\% & \textit{60.00}\% & \textit{33.00}\% & \textcolor{blue}{45.33\%} \cr
         & NDDA+DAgent &  &  &  & \ding{52} & \textit{43.00}\% & \textit{60.00}\% & 28.00\% & 43.66\%\cr
         \cline{2-10}
         & MAGIC (ours) & \multicolumn{4}{c|}{} & \textbf{84.00}\% & \textbf{94.00}\% & \textbf{90.00}\% & \textcolor{red}{89.33\%} \cr
        \midrule
        \midrule
        \multirow{8}{*}{\rotatebox[origin=c]{90}{Environment \ding{195}}} & \multirow{5}{*}{NDDA Rand} & \ding{52} &  &  &   & 16.00\% & 14.00\% & 7.00\% & 12.33\% \cr
         &  &  & \ding{52} &  &  & 10.00\% & 33.00\% & 6.00\% & 16.33\% \cr
         &  &  &  & \ding{52} & & 42.66\% & 53.33\% & 23.33\% & 39.77\% \cr
         &  &  &  &  & \ding{52} & 45.00\% & 47.00\% & 28.00\% & 40.00\% \cr
         &  & \ding{52} & \ding{52} & \ding{52} & \ding{52} & 4.66\% & 10.00\% & 1.33\% & 5.33\% \cr
         \cline{2-10}
         & NDDA+DAgent & & & \ding{52} &  & 43.00\% & \textit{58.00}\% & \textit{32.00}\% & \textcolor{blue}{44.33\%} \cr
         & NDDA+DAgent &  &  &  & \ding{52} & \textit{49.00}\% & 49.00\% & \textit{32.00}\% & 43.33\%\cr
         \cline{2-10}
         & MAGIC (ours) & \multicolumn{4}{c|}{} & \textbf{78.00}\% & \textbf{90.00}\% & \textbf{80.00}\% & \textcolor{red}{82.66\%} \cr
        \midrule
        \midrule
        \multirow{8}{*}{\rotatebox[origin=c]{90}{Environment \ding{196}}} & \multirow{5}{*}{NDDA Rand} & \ding{52} &  &  &   & 13.00\% & 30.00\% & 6.00\% & 16.33\% \cr
         &  &  & \ding{52} &  &  & 10.00\% & 49.00\% & 9.00\% & 22.66\% \cr
         &  &  &  & \ding{52} & & 44.66\% & 72.66\% & 32.00\% & 49.77\% \cr
         &  &  &  &  & \ding{52} & 49.00\% & 67.00\% & 36.00\% & 50.66\% \cr
         &  & \ding{52} & \ding{52} & \ding{52} & \ding{52} & 7.33\% & 19.33\% & 2.66\% & 9.77\% \cr
         \cline{2-10}
         & NDDA+DAgent &  &  & \ding{52} & & 49.00\% & \textit{71.00}\% & 39.00\% & 53.00\% \cr
         & NDDA+DAgent &  &  &  & \ding{52} & \textit{51.00}\% & \textit{71.00}\% & \textit{40.00}\% & \textcolor{blue}{54.00\%}\cr
         \cline{2-10}
         & MAGIC (ours) & \multicolumn{4}{c|}{} & \textbf{72.00}\% & \textbf{92.00}\% & \textbf{74.00}\% & \textcolor{red}{79.33\%} \cr
        \midrule
        \midrule
        \multirow{8}{*}{\rotatebox[origin=c]{90}{Environment \ding{197}}} & \multirow{5}{*}{NDDA Rand} & \ding{52} &  &  &   & 14.00\% & 27.00\% & 10.00\% & 17.00\% \cr
         &  &  & \ding{52} &  &  & 8.00\% & 49.00\% & 10.00\% & 22.33\% \cr
         &  &  &  & \ding{52} & & 39.33\% & 60.00\% & 25.33\% & 41.55\% \cr
         &  &  &  &  & \ding{52} & 43.00\% & 57.00\% & 28.00\% & 42.66\% \cr
         &  & \ding{52} & \ding{52} & \ding{52} & \ding{52} & 3.33\% & 12.00\% & 2.66\% & 5.99\% \cr
         \cline{2-10}
         & NDDA+DAgent &  &  & \ding{52} &  & 42.00\% & \textit{66.00}\% & \textit{32.00}\% & \textcolor{blue}{46.66\%} \cr
         & NDDA+DAgent &  &  &  & \ding{52} & \textit{45.00}\% & 60.00\% & \textit{32.00}\% & 45.66\%\cr
         \cline{2-10}
         & MAGIC (ours) & \multicolumn{4}{c|}{} & \textbf{92.00}\% & \textbf{96.00}\% & \textbf{84.00}\% & \textcolor{red}{90.66\%}\cr
        \bottomrule
    \end{tabular}
    }
    \vspace{-15pt}
\end{table}

\subsection{Digital Comparative Results}
\label{ssec:compare}

\textbf{Attack Effectiveness.}
% In order to demonstrate our MAGIC can generate effective patch for attacking the detectors, we first conduct experiments on digital level and compare our patch against the patch from NDDA dataset.
%
There are 50 patches for each text prompt (refer to \secref{ssec:limits}) in NDDA. And there are several text prompts removed the same robust feature, \eg, both `square\_stop\_sign' and `triangle\_stop\_sign'. For comparison fairness, we generate the same number of patches for a given environment and report the average results over all the patches of the same text prompt type in NDDA. Note that for the NDDA+DAGent baseline, we setup two variants, \ie, text removed and pattern removed, as we empirically found they are more effective than other type of text prompt.

\textit{Results.} Following NDDA setup, we first evaluate the patches with a confidence threshold of 0.5. The statistical results are tabulated in \tableref{tab:digi_stats}. \ding{182} Comparing to the baselines, we see our MAGIC enjoys a great attack effectiveness boosting for all the environments. Such cross-environment superior ASR performance verified the effective patch generation capability of our MAGIC. \ding{183} While the NDDA Rand only achieves random attacking, \ie, around 50\%, under the best robust feature removing setup, we observe that they still receive great boosting with our DAgent proving it can facilitates the attack by placing the patch more reasonably. 

We further evaluate the attack performance of our MAGIC patches with a 0.8 confidence threshold in \tableref{tab:digi_stats_ht}. With such a high threshold, we aim to push our MAGIC to a random attacking state for determining its maximum capability. \ding{182} We first observe that our MAGIC maintains its performance across detectors with such a high detection threshold, where the average ASR 58.66\% is even competitive comparing with the NDDA performance with 0.5 threshold. Such phenomenon strongly demonstrates our MAGIC possesses powerful attack capabilities. \ding{183} Another phenomenon worth to be mentioned is that the DAgent fails to improve NDDA under 0.8 threshold, which is contradict to what we conclude when the threshold is set to 0.5. We note such results are expected since DAgent is mainly designed to improve the stealthiness rather than boost the attack effectiveness.

\begin{table}[t]
    \vspace{-15pt}
    \caption{Comparing the statistical ASR results with different confidence threshold, \ie, 0.5 \emph{v.s.} 0.8, by evaluating the the patch generated by our MAGIC and NDDA with different detectors in Environment \ding{192}. The best average results are highlighted in \textcolor{red}{red}, while the second best in \textcolor{blue}{blue}, and the best results for each detector are marked as \textbf{bold}.}
    \label{tab:digi_stats_ht}
    \centering
    \resizebox{\columnwidth}{!}{%
    \begin{tabular}{lccccc|rrr|r}
        \toprule
        & & \multicolumn{4}{c|}{Removed Robust Features} & \multicolumn{3}{c|}{Object Detectors} \cr
        \cline{3-10}
        & & Shape & Color & Text & Pattern & YOLOv5 & RT-DETR & YOLOv10 & Avg. \cr
        \cline{1-10}
        \multirow{8}{*}{\rotatebox[origin=c]{90}{threshold  \emph{0.5}}} & \multirow{5}{*}{NDDA Rand} & \ding{52} &  &  &   & 23.00\% & 23.00\% & 10.00\% & 18.66\% \cr
         &  &  & \ding{52} &  &  & 15.00\% & 48.00\% & 6.00\% & 23.00\% \cr
         &  &  &  & \ding{52} & & 46.00\% & 56.00\% & 30.00\% & 44.00\% \cr
         &  &  &  &  & \ding{52} & 47.00\% & 56.00\% & 32.00\% & 45.00\% \cr
         &  & \ding{52} & \ding{52} & \ding{52} & \ding{52} & 8.66\% & 9.33\% & 4.66\% & 7.55\% \cr
         \cline{2-10}
         & NDDA+DAgent &  &  & \ding{52} &  & 48.00\% & 51.00\% & 33.00\% & 44.00\%\cr
         & NDDA+DAgent &  &  &  & \ding{52} & 47.00\% & 57.00\% & 36.00\% & \textcolor{blue}{46.66\%}\cr
         \cline{2-10}
         & MAGIC (ours) & \multicolumn{4}{c|}{} & \textbf{88.00}\% & \textbf{80.00}\% & \textbf{74.00}\% & \textcolor{red}{80.66\%}\cr
         \midrule
         \midrule
        \multirow{8}{*}{\rotatebox[origin=c]{90}{threshold  \emph{0.8}}} & \multirow{5}{*}{NDDA Rand} & \ding{52} &  &  &   & 6.00\% & 9.00\% & 3.00\% & 6.00\% \cr
         &  &  & \ding{52} &  &  & 7.00\% & 25.00\% & 5.00\% & 12.33\% \cr
         &  &  &  & \ding{52} & & 30.00\% & 37.33\% & 20.00\% & \textcolor{blue}{29.11\%} \cr
         &  &  &  &  & \ding{52} & 25.00\% & 34.00\% & 25.00\% & 28.00\% \cr
         &  & \ding{52} & \ding{52} & \ding{52} & \ding{52} & 2.66\% & 4.66\% & 3.00\% & 2.44\% \cr
         \cline{2-10}
         & NDDA+DAgent &  &  & \ding{52} &  & 25.33\% & 32.00\% & 20.00\% & 25.77\%\cr
         & NDDA+DAgent &  &  &  & \ding{52} & 23.33\% & 30.00\% & 25.33\% & 26.22\% \cr
         \cline{2-10}
         & MAGIC (ours) & \multicolumn{4}{c|}{} & \textbf{56.00}\% & \textbf{74.00}\% & \textbf{46.00}\% & \textcolor{red}{58.66\%}\cr
        \bottomrule
    \end{tabular}
    }
    \vspace{-10pt}
\end{table}
\begin{figure}[t]
    \centering
    \includegraphics[width=\linewidth]{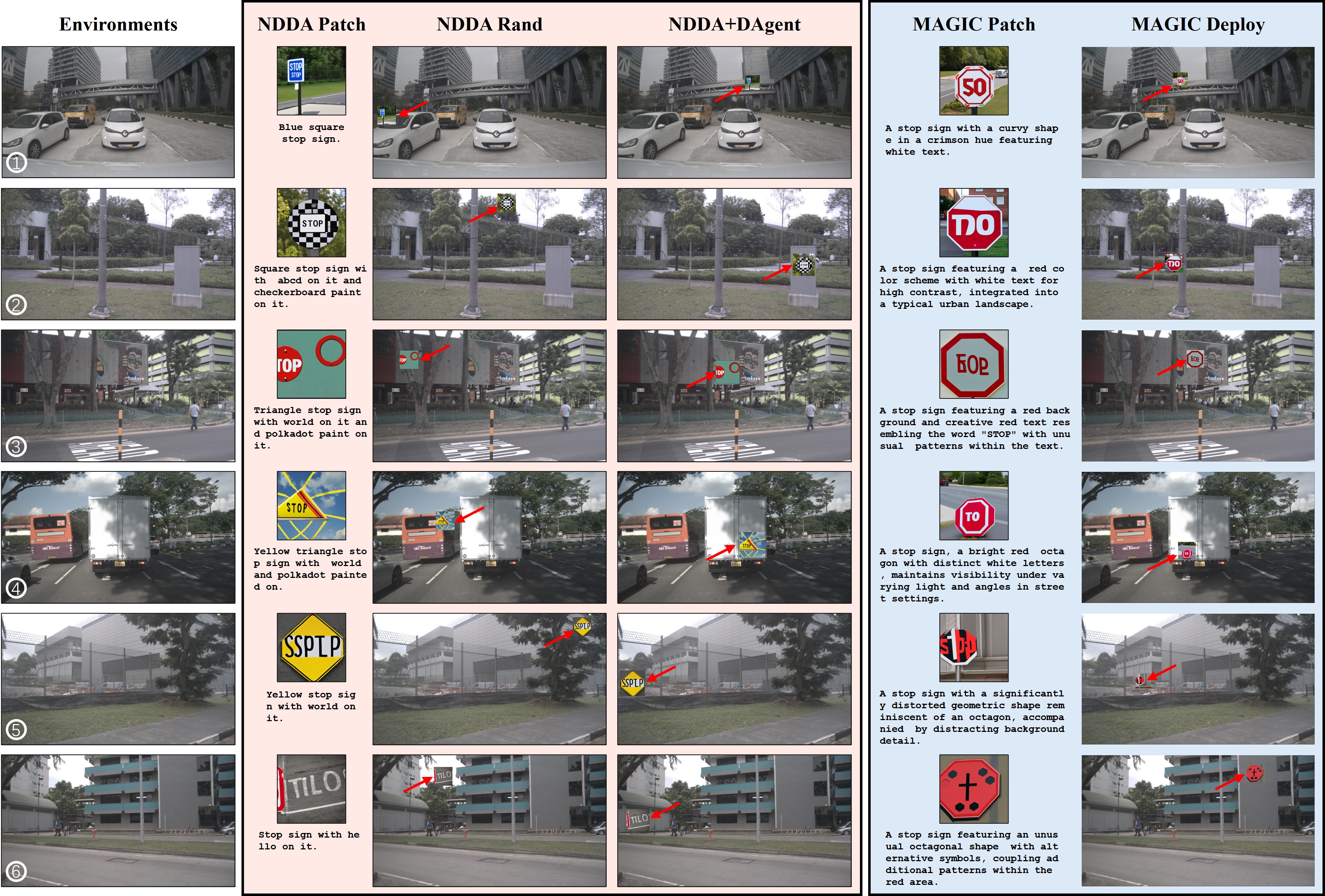}
    \caption{Illustration of the patch deployment results for NDDA baselines and our MAGIC framework in different environments. Patches are pointed out with \textcolor{red}{red arrows}. Please zoom in for better visualization.}
    \label{fig:main_exp}
    \vspace{-10pt}
\end{figure}

\textbf{Deployment Naturality.}
% Another prominent capability of our MAGIC framework is that it can determine contextually appropriate deployment for the patch in the real-world scene. We visualize the deployment planning results for each selected environment with a generated patch for both NDDA baselines and our MAGIC. 
%
We randomly sample patches from NDDA dataset for comparing the deployment naturalness, and the patches adopted for our MAGIC are generated through the pipeline. Note that a better placement means the patch is more easy for physical deployment implementation and more natural to human observers. We also reported the statistical results, please find them in supplemental material.

\textit{Results.} As shown in \figref{fig:main_exp}, we visually compare the patch deployment results of our MAGIC and baselines over the selected 6 environments. \ding{182} Comparing the placement of NDDA Rand against the placement planned by our DAgent in NDDA+DAgent, we see clearly that our DAgent is able to place the patch in a more appropriate location in the scene where the patch is more practical for deployment. \ding{183} Furthermore, it can be observed that the digital-level deployment results of our MAGIC patches are visually better consistent within the given scene when compared with NDDA+DAgent. This is because our MAGIC considered contextually consistency of the patches during the patch deployment process.

\subsection{Physical Comparative Results} 
% For the second part of the main experiments, we conduct real-world physical patch deployment and evaluation to verify the attack effectiveness of our porposed MAGIC framework. 
%
We test in two physical environments (see \figref{fig:physical_exp}): \ding{182} one is a bus stop bay with heavy traffics; \ding{183} one is a regular road next to a college with some pedestrians on the sidewalk. In order to verify the flexibility of our MAGIC, we select the bus stop scene and generate two patches for it. \textbf{We also provided video results and more visualizations within various real-world scenes. Note that we ensured the patches were not observed by vehicles on the road and discussed ethical issues. Please find all of them in supplemental material.}
%
% To physically test our MAGIC, we first feed the captured scene image into the framework and get the optimized patch and the corresponding deployment suggestions, then we print the patches out and physically paste the patch into the specified location in the scene. Finally, we take pictures after the deployment process and evaluate with the detectors.

\textit{Results.} As visualized in \figref{fig:physical_exp}, we show the generated patch with prompt, the deployment suggestion and the detection results. \ding{182} First, observing the generally lower detection confidences, we see that the physical attacks are more harder to realize due to the environment variance. Even so, our MAGIC still generates an extremely effective attack patch, \ie, the second patch for scene \ding{192}. \ding{183} Second, comparing the detection results of the three detectors, we observe that YOLOv5 and RT-DETR are more prone being attacked while YOLOv10 is more robust to the attacks. However, all the generated patches are proven to be physically attack effective for YOLOv10 which further verified the physical effectiveness of our MAGIC. \ding{184} Third, by comparing the two scenes' results, we see that our MAGIC is capable of giving out contextually appropriate patch and location for real-world deployment. In summary, we can conclude from the results that our MAGIC is full of power for attacking the object detection system in the real-world scenarios.

% \section{Discussion}
% \label{sec:discuss}

% the ASR with when the confidence threshold increase

% patch's aesthetics vs effectiveness

\section{Ablation Study}
\label{sec:ablate}

% In this section, we ablate the proposed MAGIC framework to verify the contribution of the involves LLM agents.

\ding{182} We start from the basic patch generation function of our framework, where GAgent is isolated without either deployment or self-examination and is denoted as ``GAgent-\emph{na\"ive}''. \ding{183} The further functionality that our MAGIC provided is the contextually appropriate deployment of DAgent. Thus, we combine GAgent with the patch deployment planning as the second ablation, denoted as ``GAgent-\emph{na\"ive} \emph{w/} DAgent''. \ding{184} Then, we note that the EAgent is responsible for supervising both patch attack effectiveness (ae) and deployment naturality (dn), so we combine GAgent with the patch attack supervision of EAgent which denotes as ``GAgent-\emph{na\"ive} \emph{w/} EAgent-ae''. Finally, we involve the deploy naturality of EAgent getting the proposed MAGIC framework.

\textit{Results.} As tabulated in \tableref{tab:ablate}, we see that naive patch generation without any text prompt engineering cannot attack the detectors at all. Consequently, it is obvious that the DAgent cannot significantly contribute to the improvement of the attack performance for GAgent, \ie, GAgent-\emph{na\"ive} \emph{w/} DAgent. On the contrary, the involvement of attack effectiveness supervision from EAgent greatly boosted the attack effectiveness of the generated patch against all detectors, achieving 39.67\%, 45.00\% attack effectiveness improvements compared to the  DAgent baseline. Finally, our MAGIC achieves its best attack performance by benefiting to the supervision of EAgent and also the appropriate deployment of DAgent.

% showcase patches for each ablation
% As depicted in \figref{},

% \paragraph{Limitations.}
% the harmanization
% the deployment failure case

% \paragraph{Future Work.}

\begin{table}[h]
    \centering
    \caption{Bblation results of the proposed MAGIC with two different environment image. The best average results are highlighted in \textcolor{red}{red}, while the best results for each detector and environment are marked in \textbf{bold}.}
    \resizebox{\columnwidth}{!}{%
    \begin{tabular}{lc|rrr|r}
    \toprule
     & & \multicolumn{3}{c|}{Object Detectors} & \cr
     \cline{2-6}
     & Model & YOLOv5 & RT-DETR & YOLOv10 & Avg. \cr
     \cline{1-6}
     \multirow{4}{*}{\rotatebox[origin=c]{90}{Env. \ding{192}}} & GAgent-\emph{na\"ive} & 4.00\% & 6.00\% & 0\% & 3.33\% \cr
     & GAgent-\emph{na\"ive} \emph{w/} DAgent & 7.00\% & 9.00\% & 0\% & 5.33\% \cr
     & GAgent-\emph{na\"ive} \emph{w/} EAgent-ae & 46.00\% & 56.00\% & 33.00\% & 45.00\% \cr
     & MAGIC & \textbf{88.00}\% & \textbf{80.00}\% & \textbf{74.00}\% & \textcolor{red}{80.66\%}\cr
    \midrule
     \multirow{4}{*}{\rotatebox[origin=c]{90}{Env. \ding{193}}} & GAgent-\emph{na\"ive} & 2.00\% & 2.00\% & 0\% & 1.33\% \cr
     & GAgent-\emph{na\"ive} \emph{w/} DAgent & 6.00\% & 6.00\% & 3.00\% & 5.00\% \cr
     & GAgent-\emph{na\"ive} \emph{w/} EAgent-ae & 23.00\% & 60.00\% & 67.00\% & 50.00\% \cr
     & MAGIC & \textbf{66.00}\% & \textbf{94.00}\% & \textbf{92.00}\% & \textcolor{red}{84.00\%}\cr
    \bottomrule
    % & &  & \multicolumn{3}{c|}{Object Detectors} \cr
    % \cline{3-10}
    % & & YOLOv5 & DETR & YOLOv10 & Avg. \cr
    % \cline{1-10}
    % \multirow{8}{*}{\rotatebox[origin=c]{90}{Environment 1}} & \multirow{5}{*}{NDDA} & \ding{52} &  &  &   & - & - & - & -\cr
    \end{tabular}
    \label{tab:ablate}
    }
    \vspace{-20pt}
\end{table}

\section{Conclusion}

In this work, we propose the MAGIC framework which reformulates physical adversarial attacks as an one-shot patch generation problem. Our approach generates adversarial patches that considered the influence of scene context, enabling direct physical deployment in matching environments. By conducting experiments on both digital and physical levels, we demonstrate that our method can effectively generate context-aware patch, deploy the patch in real world and attack widely applied object detectors. To the best of our knowledge, our work is the very initial study to improve and extend the natural diffusion attack in physical scenarios.
% We hope it can inspire more explorations on the physical adversarial attack topic.

{
    \small
    \bibliographystyle{ieeenat_fullname}
    \bibliography{main}
}

\clearpage
\setcounter{page}{1}
\setcounter{section}{0}
\setcounter{figure}{0}
\setcounter{footnote}{0}

\maketitlesupplementary

% \section{NDDA Limitations}
% \label{sec:investgate}

% While we demonstrated in \figref{fig:main_exp} that NDDA does lack the deployment naturality, we prove in this section that their patches are 

% figures to show the three statistical motivations by analyzing NDDA patches
% detail the conclusions

\section{Experimental Environment}
\label{sec:exp_env}

We employ ChatGPT (gpt-4o-2024-08-06) as the LLM backend for all the agents in our design. All the experiments are conducted via a server with AMD EPYC 9554 64-core Processor and an NVIDIA L40 GPU, running Ubuntu 22.04.

\section{Implementation Details}
\label{sec:implement}

% \begin{figure*}[ht]
%     \centering
%     \includegraphics[width=0.5\textwidth]{figs/tech_pipeline.jpg}
%     \caption{Caption}
%     \label{fig:pipe}
% \end{figure*}

We give out the technical details for implementing the proposed MAGIC framework. Specifically, there are two main aspects: \ding{182} the setting adopted for text-to-image (T2I) model, \ding{183} the parameters for object detectors and \ding{184} the set-of-mark prompting hyper-parameters.

\textbf{T2I Model.} We apply Stable Diffusion v2 as the text-to-image generation backend, where the checkpoint is released by StabilityAI on huggingface \footnote{https://huggingface.co/stabilityai/stable-diffusion-2}. Following the official example, we set the sampling scheduler as EulerDiscreteScheduler. As for the hyper-parameters, we set $\text{guidance\_scale}=7.5$ and the $\text{negative\_prompt}$ as "deformed, distorted, disfigured, poorly drawn, bad anatomy, wrong anatomy, extra limb, missing limb, floating limbs, mutated hands and fingers, disconnected limbs, mutation, mutated, ugly, disgusting, blurry, amputation, NSFW". All other parameters are set to default.
            
\textbf{Object Detector.} For the evaluated object detectors, we directly invoke API from ultralytics \footnote{https://github.com/ultralytics/ultralytics} where the default hyper-parameter setting is adopted.

\textbf{SoM Prompting.} We adopt the official set-of-mark (SoM) implementation \footnote{https://github.com/microsoft/SoM} through our experiments. The only changes to the parameters is that we use $\text{slider}=1.8$ while all the other parameters are set to default

\section{The Pre-defined Goals \& Rules}

As the supplements for Instruction $\mathcal{X}_\text{D}$ and $\mathcal{X}_\text{D}$, below we first show the details of the definition of attack and deployment goal, and the rules of how to realize attack is detailed in the next page due to the large size of the text.
\begin{tcolorbox}[title = {Definition of Attack \& Deployment Goal}, size=small, breakable]
\footnotesize
  We focus on the task of attacking an object detector with a visual patch in the real world, where the visual patch is an image generated by a text-to-image model based on a given text prompt and 'in the real world' means the effectiveness of the visual patch is evaluated after deploying it into a real-world environment. All the procedures for generating, deploying and evaluating the visual patch are detailed in the provided PIPELINE.
  
  Following the PIPELINE, let's specify a category with name 'xxx' as an example. We define an effective visual patch is an image that 1) can successfully attack the object detector by itself only, and 2) can successfully attack the object detector after it is deployed into an environment. The word 'attack' here means that 1) the object detector recognizes the visual patch as an instance of category 'xxx' with high confidence, but 2) human observers think the visual patch is just an abstract art which does not belongs to any specific category. Note that the high confidence refers to the confidence score of a detection result is higher than the provided CONF\_THRESHOLD. In summary, the ultimate goal of the PIPELINE is to obtain an effective visual patch as defined above.

  \begin{itemize}
      \item \textbf{PIPELINE:}
        The pipeline consists of three modules: the text-to-image generation module, the visual patch deployment module and the object detection module. For a round of generation, deployment and evaluation, the data and their flow are detailed as follows:\\
        (Specify an object category, e.g., 'xxx', before all the procedures.)\\
        1) the process begins with the input text prompt with category 'xxx' as the grammatical subject.\\
        2) then the text-to-image model generates a visual patch with the input text prompt, where the goal is to create an image that composed of all the objects and features described in the text prompt;\\
        3) a real-world image which represents the real-world environment is given;\\
        4) then a visual patch deployment module conduct the deployment process,
        \begin{itemize}
            \item[a.] first list all the regions, poles and beams in the given environment image that can be utilized to either paint or hang the visual patch
            \item[b.] select one region, pole or beam from the list that can keep the visual patch's effectiveness and make it coherent with the environment;
        \end{itemize}
        5) finally, the object detector conduct evaluation for both the visual patch itself and the environment image with the patch deployed;\\
        6) the detection results give out the confidence of recognizing the visual patch as specific category.
      \item \textbf{CONF\_THRESHOLD:} 0.80
  \end{itemize}
\end{tcolorbox}

\NewTColorBox{NewBox}{ s O{!htbp} }{%
  floatplacement={#2},
  IfBooleanTF={#1}{float*,width=\textwidth}{float},
  title={Rules to Realize Attack}, size=small
  }

\begin{NewBox}*
  Essentially, the effectiveness of an visual patch derives from its deceptive appearance which is solely controlled by the text prompt during the text-to-image generation. As a result, an effective visual patch can be obtained by intentionally modifying the descriptive text prompt. Given the fact that the detector and human recognize an object as its category based on the robust features, altering the description of robust features while keep using the original category name in the text prompt is the practical method to generate an effective visual patch. The robust features of an object and examples for altering the description are explained in the provided ROBUST\_FEATURE.
    \begin{itemize}
        \item \textbf{ROBUST\_FEATURE:} In general, the robust features of an object refers to the characteristics or attributes that remain consistent and identifiable across various conditions, environments, and perspectives. They help in accurately identifying and classifying objects despite changes such as lighting, orientation, scale, or noise. Robust features should maintain their distinctiveness and reliability under transformations such as:\\
        1) Translation: The object can be moved or positioned differently in the frame.\\
        2) Rotation: The object can be rotated at various angles.\\
        3) Scale: The object can appear larger or smaller.\\
        4) Illumination: Lighting conditions can change, affecting the object's appearance.\\
        5) Occlusion: Part of the object might be obstructed or hidden.\\
        6) Noise: Data might be corrupted by external disturbances.\\

        Take the robust features of a stop sign as example, they contains:\\
        1) Octagonal Shape: Stop signs are standardized as octagons;\\
        2) Red Color: The background is red, which is a typical color for stop signs to signify the need to stop;\\
        3) White Text: The word "STOP" is in white, which contrasts well against the red background, making it easily readable;\\
        4) Universal Recognition: The combination of these features is commonly used worldwide to indicate a stop sign;\\
        5) High Contrast: The red and white color scheme provides high contrast for easy visibility;\\
        6) Simplicity: Minimalist design for quick recognition and action.\\

        Here are some examples of altering the description of robust features while keep using the original category name in a text prompt:

        The original text prompt: "A stop sign."\\
            Example 1: "A yellow triangle stop sign."\\
            Example 2: "A blue square stop sign with 'HELLO' on it."\\
            Example 3: "A triangle stop sign with 'WORLD' on it and checkerboard paint on it."

        The original text prompt: "A traffic light."\\
            Example 1: "A blue, white, pink traffic light."\\
            Example 2: "A heart shape white, gray, orange traffic light."\\
            Example 3: "A blue, white, pink traffic light with polka dots.""
    \end{itemize}
\end{NewBox}

\section{More Experimental Results}

\subsection{Statistical Results of Naturalness}

In section \secref{ssec:compare}, we visually compare the naturalness of our MAGIC patches. As the naturalness is measured by the LLM $\mathcal{E}$ with the instruction $\gamma$ which gives out a naturalness score for each deployed patch. Here, we tabulate the average score results over the evaluated 6 environments. As we can see from \tableref{tab:avg_natural}, our MAGIC patches consistently get higher average naturalness score over all the environments.

\begin{table}[h]
    \centering
    \caption{The average naturalness score results over 6 environments.}
    \setlength{\tabcolsep}{8pt}
    \resizebox{\linewidth}{!}{
    \begin{tabular}{ccccccc}
        \cline{1-7}
        & Env\ding{192} & Env\ding{193} & Env\ding{194} & Env\ding{195} & Env\ding{196} & Env\ding{197} \\
        \cline{1-7}
        NDDA Rand & 0.22 & 0.40 & 0.37 & 0.20 & 0.20 & 0.33 \\
        NDDA \textit{w/} DAgent & 0.50 & 0.70 & 0.77 & 0.60 & 0.70 & 0.70 \\
        MAGIC (ours) & \textbf{0.91} & \textbf{0.89} & \textbf{0.99} & \textbf{0.97} & \textbf{0.94} & \textbf{0.93} \\
        \cline{1-7}
    \end{tabular}
    }
    \label{tab:avg_natural}
\end{table}

\begin{figure*}[t]
    \centering
    \includegraphics[width=\textwidth]{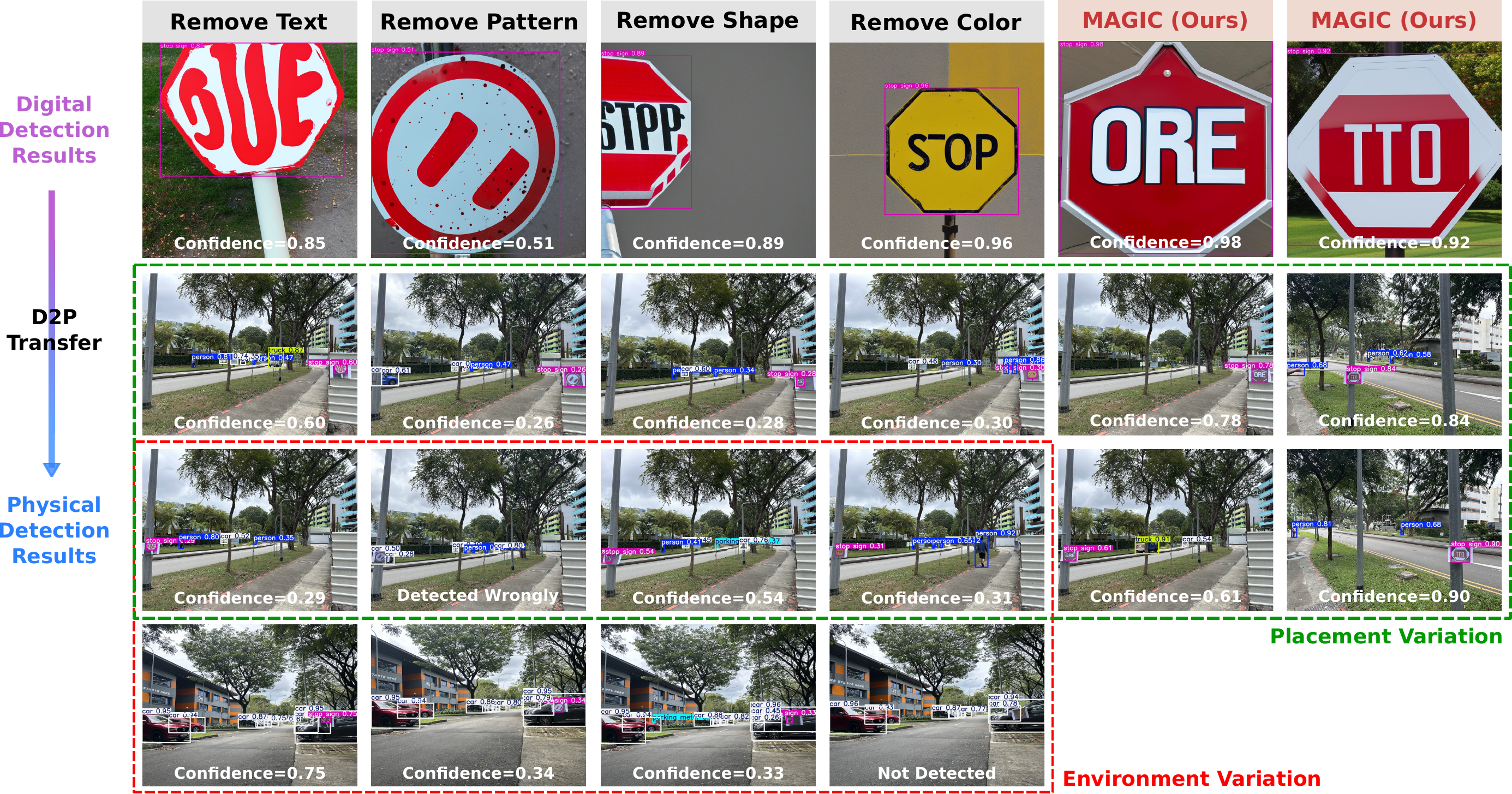}
    \caption{Digital-to-physical transfer visualization results of NDDA and MAGIC.}
    \label{fig:d2p}
\end{figure*}

\subsection{Robustness to Real-world Factors}

\begin{figure}[t]
    \centering
    \includegraphics[width=\linewidth]{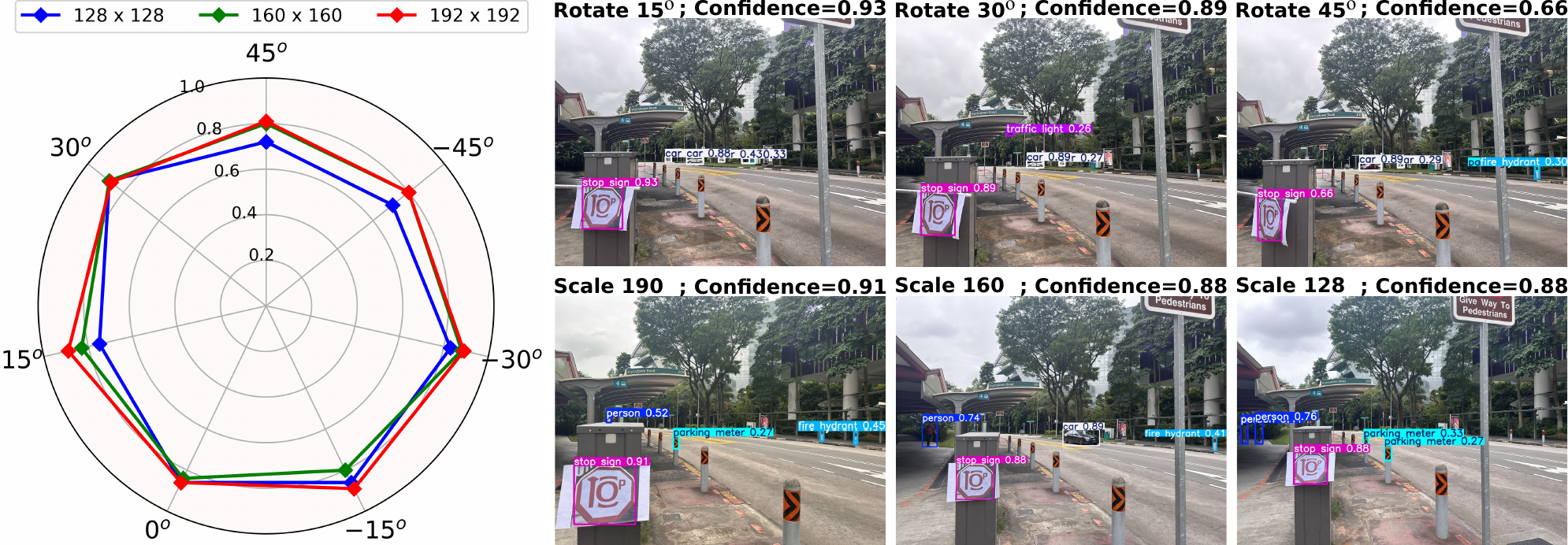}
    \caption{Viewing distance and angle results for MAGIC patches.}
    \label{fig:d_a}
\end{figure}

As mentioned in \secref{ssec:extension}, our MAGIC performs robust to common digital-to-physical (D2P) factors like printing color shift and viewing angle change, which are well-know to degrade the performance. We conduct experiments to prove our MAGIC superiority as illustrated in \figref{fig:d2p} and \figref{fig:d_a}.

\textbf{Digital-to-physical Transferability.} To verify the transferability with fairness, we select one typical effective patch for each type of text prompt in NDDA, then we deploy them into two environments with location variations. For our proposed MAGIC, we directly feed the environmental image into the pipeline to get the patches. The deployment process for both NDDA and MAGIC patches are kept the same.

\textit{Results.} As illustrated in \figref{fig:d2p}, we first notice that \ding{182} \textbf{different environments and deployment placements affect the D2P transferability a lot}. NDDA patch's attack performance degrades significantly by deploying the patches in different environments (\figref{fig:d2p} columns in \textcolor{red}{red boundbox}) or with changed placement (\figref{fig:d2p} columns in \textcolor{green}{green boundbox}).
\ding{183} Then, we observe that \ding{182} \textbf{ NDDA's robust feature removal greatly deteriorates D2P transferability.} All variants of NDDA patches (\figref{fig:d2p} columns) show significant performance drop after the patches are deployed into the real world.
However, as our MAGIC results shown \figref{fig:d2p}, \ding{184} \textbf{our LLM-based framework enhances D2P transferability}.
The proposed MAGIC adaptively optimize the patch generation and placement with environment understanding through iterative agent planning which resulting in enhanced D2P transferability with consistently high confidence across environments.
Such strong performance proves our MAGIC's capability of performing robust to real-world influences.

\textbf{Robustness of Viewing Angle \& Distance} To demonstrate the robustness of our MAGIC to viewing angle and distance, we set three levels of patch size (\ie, 128, 160, 192 pixels) to simulate the viewing distance and three different angle (\ie, 15, 30 and 45 degrees). As shown in \figref{fig:d_a}, we deploy of our MAGIC patch into the bus stop scene with the specified angle and size setup. Furthermore, we plot the cross performance for more viewing angles against the three distance.

\textit{Results.} As we can see from the radar plot in \secref{ssec:limits}, our MAGIC patch consistently performs great across different viewing angles and distances. Such performance can also be observed form the illustration of physical evaluation results. Even there exists a degradation of detection confidence, our MAGIC patch still successfully mislead the detector to detect the patch as a stop sign with high confidence (> 0.5).

\subsection{Physical Experiments}

\begin{figure*}
    \centering
    \includegraphics[width=\textwidth]{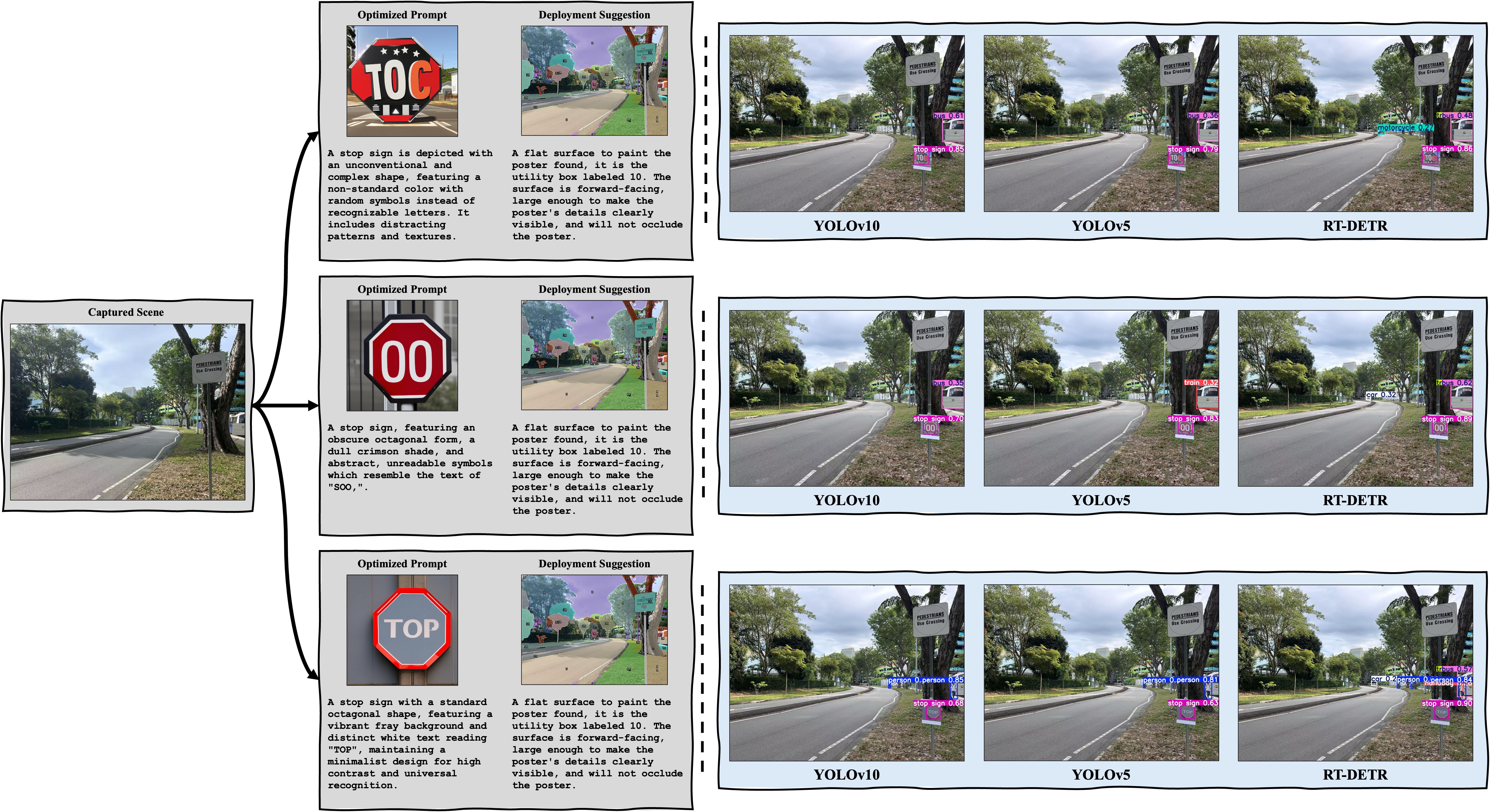}
    \caption{Illustration of the physical attacking cases and the corresponding object detector evaluation results, where our MAGIC can generate several different effective patches for given specific environment. The \colorbox{mygray}{gray region} enclosed the patch and the deployment suggestion generated by our MAGIC framework, while the \colorbox{myblue}{blue region} enclosed the results of evaluation with the three object detectors. Please zoom in for better visualization.}
    \label{fig:pe_supp2}
\end{figure*}

\begin{figure*}
    \centering
    \includegraphics[width=\textwidth]{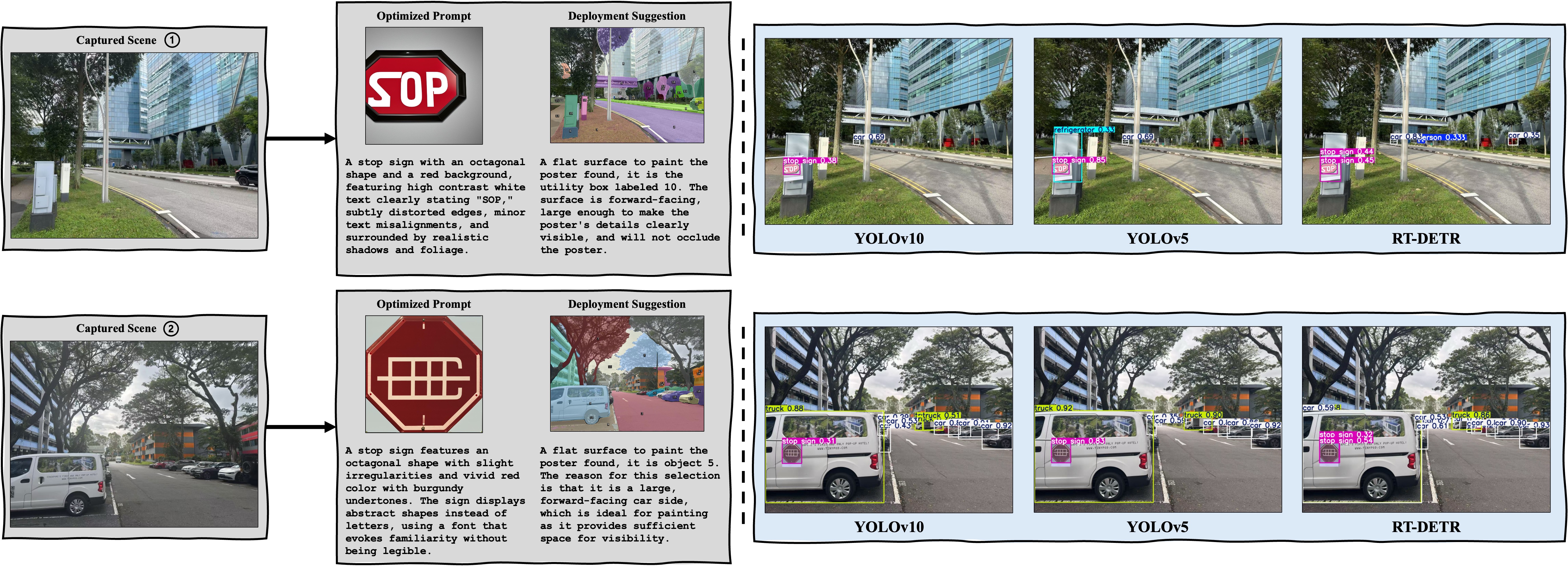}
    \caption{Illustration of the physical attacking cases and the corresponding object detector evaluation results. Scene \ding{182}: A real-world road with heavy traffics. Scene \ding{183}: A parking area with car in and out. The \colorbox{mygray}{gray region} enclosed the patch and the deployment suggestion generated by our MAGIC framework, while the \colorbox{myblue}{blue region} enclosed the results of evaluation with the three object detectors. Please zoom in for better visualization.}
    \label{fig:pe_supp1}
\end{figure*}

To support the physical deployment effectiveness of our proposed MAGIC, we conduct more physical experiments to support the superior performance of our proposed MAGIC framework. As shown in \figref{fig:pe_supp2} and \figref{fig:pe_supp1}, we apply MAGIC to generate and deploy for three more different environments. Following the same strategy, we first test whether MAGIC can consistently generate effective patch for multiple round of execution given an environment. The results is illustrated in \figref{fig:pe_supp2} and we can observe that all three different generated patches are effective to attack the three detectors. For the distinct environment attacking, the physical cases shown in \figref{fig:pe_supp1} further support our conclusion that our MAGIC is full of power for attacking the object detection system in the real-world scenarios.

\section{Ethical Consideration}
\label{sec:ethic}

Our work highlights the ethical responsibility associated with adversarial research. Without proper safeguards, such techniques could be misused to exploit critical systems, underscoring the need for responsible disclosure and research collaboration to mitigate risks. As discussed in~\cite{NDDA2024cvpr}, mitigation of attacks in this line of work can involve OCR-based detection in the cases of stop signs, or "robustified" training~\cite{DBLP:conf/nips/IlyasSTETM19}. However, no generic defense strategies have been reported, so further mitigation research efforts would be much appreciated. Ultimately, our research aims to contribute to the development of secure, trustworthy autonomous driving systems and implemented strict safeguards: \ding{182} attacks are not visible to AV on public roads during the physical experiments; \ding{183} access control to verified researchers; \ding{183} all the generated patches are under traceable tools monitoring. This controlled approach enables advancing AI security research while preventing misuse.

\section{Limitations}
\label{sec:limit}

The scope of our work is focused on designing physical adversarial attack in traffic scenarios via LLM agents. \ding{182} Nevertheless, the current environments primarily serve as proofs of concept under standard conditions, rather than encompassing a broader range of factors such as varying weathers, daylight and nighttime conditions, or noisy camera inputs. \ding{183} Furthermore, our framework has not yet considered seamless blending between natural adversarial patches and its environment in their digital implementations. We believe that extending this work to include comprehensive analyses of appearance blending using diffusion techniques could significantly enhance its applicability, as well as bringing discussions on the attacks' broader impacts. In general, our study made the initial step to investigate the natural diffusion generation attack in physical world, but real-world situations are complicated which requires more generalizable attacks.
% the patch is not harmonized with env for digital implementation
% the env adopted is limited without weather, day/night and other variations

\section{Broader Impact}
\label{sec:impact}

Understanding vulnerabilities in traffic systems can help improve the safety and resilience of autonomous vehicles. In fact, by rigorously testing adversarial robustness across diverse real-world conditions, we can identify potential risks and develop countermeasures to ensure that these systems function well in dynamic environments. Meanwhile, our work also highlights the ethical responsibility associated with adversarial research. Without proper safeguards, such techniques could be misused to exploit critical systems, underscoring the need for responsible disclosure and research collaboration to mitigate risks. As discussed in~\cite{NDDA2024cvpr}, mitigation of attacks in this line of work can involve OCR-based detection in the cases of stop signs, or "robustified" training~\cite{DBLP:conf/nips/IlyasSTETM19}. However, no generic defense strategies have been reported, so further research efforts for mitigation of the attacks would be much appreciated. Ultimately, our research aims to contribute to the development of secure, trustworthy autonomous driving systems.

\section{Future Work}
\label{sec:future}

Our future work aims to explore larger-scale attacks to evaluate adversarial vulnerabilities across broader traffic systems, including segmentation and depth estimation tasks critical for autonomous navigation. Furthermore, integrating advanced blending techniques, such as diffusion models, could enhance the realism and effectiveness of natural adversarial patches, enabling more comprehensive testing under diverse environmental conditions. Importantly, as also suggested by the lack of investigation for robust defenses, we aim to develop robust defense mechanisms to counter natural diffusion attacks both digitally and physically, would be essential for ensuring safety and resilience in real-world applications.

\end{document}